\documentclass[12pt]{article}

\usepackage{arxiv}        

\usepackage{graphicx}
\usepackage{enumitem}
\usepackage{amsmath}
\usepackage{amssymb}
\usepackage{cite}
\usepackage{subfigure}
\usepackage{algorithm}
\usepackage{algorithmic}
\usepackage{bm}
\usepackage{bbm}
\usepackage{url}
\usepackage{mathtools} 
\mathtoolsset{
  showonlyrefs,
  showmanualtags,
}

\allowdisplaybreaks  %

\newcommand{\dd}{\mathrm{d}}

\newcommand{\bE}{\mathbb{E}}
\newcommand{\bR}{\mathbb{R}}

\newcommand{\cE}{\mathcal{E}}

\newcommand{\cX}{\mathcal{X}}

\newcommand{\cB}{\mathcal{B}}

\newcommand{\cD}{\mathcal{D}}
\newcommand{\cF}{\mathcal{F}}

\newcommand{\cL}{\mathcal{L}}
\newcommand{\cM}{\mathcal{M}}

\newcommand{\cP}{\mathcal{P}}

\newcommand{\cS}{\mathcal{S}}

\newcommand{\argmax}{\mathop{\rm arg~max}\limits}
\newcommand{\argmin}{\mathop{\rm arg~min}\limits}

\newcommand{\minimize}{\mathop{\rm minimize}\limits}
\newcommand{\lan}{\langle}
\newcommand{\ran}{\rangle}

\newtheorem{defi}{Definition}

\newcommand{\appropto}{\mathrel{\vcenter{
  \offinterlineskip\halign{\hfil$##$\cr
    \propto\cr\noalign{\kern2pt}\sim\cr\noalign{\kern-2pt}}}}}

\newcommand{\new}{\mathrm{new}}

\title{Geometry of EM and related iterative algorithms
}

\author{Hideitsu Hino\\
            The Institute of Statistical Mathematics, Tokyo 190-8565, Japan\\
            RIKEN AIP, Tokyo 103-0027, Japan\\
  \texttt{hino@ism.ac.jp}
\And
        Shotaro Akaho\\
        The National Institute of Advanced Industrial Science and Technology, Ibaraki 305-8568, Japan
        \And
        Noboru Murata\\
        Waseda University, Tokyo 169-8555, Japan
}

\begin{document}
\maketitle

\begin{abstract}
The Expectation--Maximization (EM) algorithm is a simple meta-algorithm that has been used for many years as a methodology for statistical inference when there are missing measurements in the observed data or when the data is composed of observables and unobservables. Its general properties are well studied, and also, there are countless ways to apply it to individual problems. In this paper, we introduce the $em$ algorithm, an information geometric formulation of the EM algorithm, and its extensions and applications to various problems. Specifically, we will see that it is possible to formulate an outlier-robust inference algorithm, an algorithm for calculating channel capacity, parameter estimation methods on probability simplex, particular multivariate analysis methods such as principal component analysis in a space of probability models and modal regression, matrix factorization, and learning generative models, which have recently attracted attention in deep learning, from the geometric perspective provided by Amari.
\keywords{Information Geometry, EM Algorithm, $em$ Algorithm, Bregman Divergence, Information Theory, Robust Statistics, Generative Models}
\end{abstract}

\section{Introduction}
The Expectation--Maximization (EM) algorithm is a maximum likelihood estimation algorithm for missing observations proposed in~\cite{DEMP1977}. 
The EM algorithm consists of an E-step that fills in missing parts of the observed data to generate pseudo-complete data and an M-step that maximizes the likelihood function for the complete data. The E-step can be described using the sufficient statistics of the assumed statistical model, while the M-step specifically solves the likelihood equation in the framework of complete data.
The EM algorithm is well established as a general-purpose numerical solution for maximum likelihood estimation of missing observations.
The regularity conditions for convergence and the  convergence conditions for the sequence of log-likelihood function values and parameter estimates generated by the EM algorithm were investigated in~\cite{10.1214/aos/1176346060}, and a convergence rate and its estimation method of the algorithm were also developed~\cite{10.2307/2337198}.
In~\cite{1574231875723835008}, Csiszar and Tusnady studied sufficient conditions for the convergence of algorithms that find the shortest distance between two sets by iterations involving the EM algorithm and gave examples of calculating the channel capacity, rate distortion function, and portfolio optimization. 

Statistical properties and other variants of the EM algorithm are summarized in, for example,~\cite{BA85746989}.
Even in recent years, various novel theoretical results on the EM algorithm have been discovered. For example, in \cite{Balakrishnan2017}, a theoretical foundation for quantifying the convergence of the EM algorithm within a statistical precision of a global optimum was developed, while in~\cite{DBLP:conf/aistats/KwonHC21}, a strong theoretical guarantee of the EM algorithm applied to a mixture linear model was established. 
It is also widely used in applications such as machine learning, information theory, imaging~\cite{doi:10.1177/096228029700600105}, epidemiology~\cite{McLachlan1997TheIO,doi:10.1177/096228029700600103}, psychology~\cite{Enders2003UsingTE}, privacy~\cite{DBLP:journals/tifs/MurakamiKH17,DBLP:journals/popets/MurakamiHS18}, neuroscience~\cite{DBLP:journals/nn/IwasakiHTAM18}, and economics~\cite{RUUD1991305}, and is being extended in each of these fields. For example, for estimating the parameters of the hidden Markov model, the Baum--Welch algorithm~\cite{10.1214/aoms/1177699147} which is nothing but an instance of the EM algorithm, is widely used.

Information geometry is a framework for analyzing the statistical manifold equipped with the Fisher metric and a pair of affine connections with the methodology of differential geometry~\cite{amari2000methods}. The information geometry makes it possible to understand the mechanisms and behavior of statistical estimation and machine learning in relation to the structure of the space of probability distributions. The geometric view has yielded a variety of results. For example, it has been used to clarify the relationship between predictive distribution and curvature in Bayesian statistics~\cite{10.2307/2337602}. In semiparametric inference, it is used to decompose the parameter of interest and the nuisance parameter by orthogonal foliation~\cite{bj/1178291931}. It offers an orthogonal decomposition of hierarchical statistical models such as a nested stochastic dependence among a number of random variables such as a higher order Markov chain~\cite{930911}. Ensemble methods in machine learning, such as Bagging~\cite{Breiman1996} and Boosting~\cite{FREUND1997119}, are also investigated from the viewpoint of information geometry. The Bagging predictor is analyzed in~\cite{Baggins2004}, and it is shown that bootstrap predictive distributions are equivalent to Bayesian predictive distributions in the second-order expansion. The geometric structure of the Boosting algorithm has been elucidated in~\cite{NIPS2001_71e09b16} by identifying a classification problem with an estimation problem of conditional probability. In~\cite{DBLP:journals/neco/MurataTKE04}, the inference procedure in Boosting algorithms was extended by considering the class of  $U$-divergence, which is an extension of a standard Kullback--Leibler divergence, and the robustness of the information geometrically extended Boosting algorithms is investigated in~\cite{DBLP:journals/neco/TakenouchiEMK08}.

The EM algorithm was characterized from a geometric perspective in~\cite{AMARI19951379}. Because of this pioneering work, the usefulness of considering iterative algorithms from a geometric point of view is now widely known, and inference algorithms in various aspects have been analyzed in the information geometric framework. In this paper, we provide an overview of EM-like algorithms with iterative structures from a geometric point of view; since the EM algorithm and its applications are very broad, we aim to provide a concise survey focusing on the geometric point of view. The rest of the paper is organized as follows. In Sections~2 and 3, the EM algorithm and the element of information geometry are presented. Section~4 introduces the $em$ algorithm, the information geometric version of the EM algorithm. From Sections~5 to 8, various iterative algorithms are considered from the viewpoint of information geometry. In Section~5, geometrical analysis of an algorithm for calculating the capacity of a memoryless communication channel is presented. Section~6 deals with parameter estimation problems for statistical models with special structures. Section~7 considers the situation that a distribution is regarded as a datum, and a principled framework for dealing with distributional data is introduced. Section~8 shows an attempt to formulate generative model learning from a geometric manner. Section~9 is devoted to conclusions.

\section{EM algorithm}
The EM algorithm is a method of performing maximum likelihood estimation by simple iterative computation for problems where a part of the random variable is unobservable for some reason. The EM algorithm can be applied to the parameter estimation of mixture models by treating the unknown information concerning which distribution the data were observed from as a missing variable.
In this section, we introduce the symbols and describe  the problem setup through a description of the EM algorithm.

Let $X$ be a random variable and $x$ be its realization. Let $Z$ be the hidden variable. In other words,  we consider the situation that a part of a random vector is observed while the rest cannot be observed. The problem is to determine the parameter $\theta$ of the statistical model $p(x,z;\theta)$ from only the observations of $X$, where its marginal distribution is given by
\begin{align}
    p(x ; \theta) = \int p(x,z; \theta) \dd z
\end{align}
Taking the logarithm of both sides gives
\begin{align}
    \log p(x;\theta) =& \log \int p(x,z;\theta) \dd z
\shortintertext{but the logarithm of the summation (integration) is intractable in general; hence we take the variational lower bound as}
    =& \log \int q(z)\frac{p(x,z;\theta)}{q(z)} \dd z \\
    \geq & \int q(z) \log \frac{p(x,z;\theta)}{q(z)} \dd z \\
    =:& \cL (q,\theta)
\end{align}
where the inequality comes from Jensen's inequality. Note that
\begin{align}
    \log p(x;\theta) - \int q(z) \log \frac{p(x,z;\theta)}{q(z)} \dd z 
    =& \int q(z) \log p(x;\theta) \dd z - 
    \int q(z) \log 
    \frac{
    p(z|x;\theta) p(x;\theta)
    }{ q(z)} \dd z \\
    =&
    \int q(z) \log \frac{q(z)}{p(z|x;\theta)} \dd z
    = D(q(z) , p(z|x;\theta)),
\end{align}
where 
\begin{align}
    D(f , g) = \int \left( f(x) \log \frac{f(x)}{g(x)} \right) \dd x
\end{align}
is the Kullback--Leibler (KL) divergence. 

Suppose a set of observation $\{x_i\}_{i=1}^{n}$ is given.
Then, starting from an initial parameter $\theta_0$ and $t=0$, the EM algorithm is the following iterative procedure composed of the E- and M-steps.
\begin{description}
    \item[E-step: ] Maximize the variational lower bound $\cL(q,\theta_t) 
    = \int q(z) \log \frac{p(x,z;\theta_t)}{q(z)} \dd z$ w.r.t. $q$. Namely, 
    \begin{align}
        \minimize_{q(z)} \; D(q(z) , p(z|x,\theta_t)),
    \end{align}
    which is achieved by setting $q(z)$ to be the estimated posterior as $q(z) = p(z|x;\theta_{t})$, and calculate the Q-function as
    \begin{align}
    Q(\theta,\theta_{t}) := \frac{1}{n} \sum_{i=1}^{n} \int p(z|x_i;\theta_{t}) \log p(x_i,z;\theta) \dd z + const.
    \end{align}
    \item[M-step: ]Maximize $\cL(q,\theta)$ with respect to $\theta$ and update $\theta_t$
    \begin{align}
        \theta_{t+1} = \argmax_{\theta} \;
        Q(\theta,\theta_{t}).
    \end{align}
\end{description}
The EM algorithm is also used for MAP estimation
\begin{align}
    \mbox{maximize} \; \log p(\theta|x)
\end{align}
when a prior distribution $p(\theta)$ is given. The posterior distribution is
\begin{align}
    \log p(\theta | x) =& 
    \log 
    \frac{p(x|\theta) p(\theta)}{p(x)} = 
    \log 
    \frac{p(\theta)}{p(x)} \int p(x,z|\theta) \dd z \\
    =& 
    \log \frac{p(\theta)}{p(x)} \int q(z) \frac{p(x,z|\theta)}{q(z)} \dd z \\
    =& 
    \log p(\theta) - \log p(x) + \log \int q(z) \frac{p(x,z|\theta)}{q(z)} \dd z \\
    \geq & \log p(\theta) - \log p(x) + \int q(z) \log \frac{p(x,z|\theta)}{q(z)} \dd z \\
    =& 
    \cL(q,\theta) + \log p(\theta) - \log p(x) = \cL^{\prime}(q,\theta),
\end{align}
and we have
\begin{equation}
    \log p(\theta|x) = \cL^{\prime}(q,\theta)
    +
    D(q(z) , p(z|x,\theta)).
\end{equation}
The E-step for MAP estimation is the same as the standard E-step, while the M-step for MAP estimation maximizes
\begin{equation}
    Q(\theta, \theta_{t}) + \log p(\theta).
\end{equation}

\section{Information geometry}
Let us consider the space of positive finite measures over $x \in \cX$, where $\cX$ is a space of input variables, under a carrier measure $\Lambda (x)$
\begin{equation}
    \cF = 
    \left\{
    m(x) \; \middle| \; m: \cX \to \bR_{+}, \; \int_{x \in \cX} m(x) \dd \Lambda (x) < \infty,
    \right\}
\end{equation}
and the space of probability densities as a subspace of $\cF$
\begin{equation}
    \cS =
    \left\{
    m(x) \; \middle| \; m : \cX \to \bR_{+}, \; \int_{x \in \cX} m(x) \dd \Lambda (x) =1
    \right\} \subset \cF.
\end{equation}
We restate the KL divergence with more generality as
\begin{align}
    D(f, g) = \int f \log \frac{f}{g} \dd \Lambda.
\end{align}
The integral with respect to the measure $\Lambda$ should read summation when we consider discrete variables. 
When the KL divergence is adopted for measuring a statistical distance between distributions, the $m$-geodesic and the $e$-geodesic play the most important roles. The $m$-geodesic is defined as a set of interior points between two distributions $p(x)$ and $q(x)$:
\begin{equation}
    r(x;t) = (1-t) p(x) +  t q(x), \quad t \in (0,1).
\end{equation}
The $e$-geodesic is defined as a set of interior points between $p(x)$ and $q(x)$ in the sense of the logarithmic representation:
\begin{equation}
    \log r(x;t) = (1-t) \log p(x) + t \log q(x) + a(t), \quad t \in (0,1)
\end{equation}
where $a(t)$ is the normalization constant to make $r(x;t)$ a probability function and is defined by
\begin{equation}
    a(t) = \log \int p(x)^{1-t} q(x)^t \dd x.
\end{equation}

Let $K$ be a submanifold of $\cS$ and $p \in \cS$. We call $\hat{p}$ an $m$-projection of $p$ onto $K$ when the $m$-geodesic connecting $p$ and $\hat{p}$ is orthogonal to $K$ with respect to the Fisher metric $g$ at $\hat{p}$. Also, we call $\hat{p}$ an $e$-projection of $p$ onto $K$ when the $e$-geodesic connecting $p$ and $\hat{p}$ is orthogonal to $K$ at $\hat{p}$.

In information geometry~\cite{amari2000methods}, a manifold that consists of statistical models is called a model manifold and is denoted by $\cM$. One of the representative parametric models is the exponential family 
\begin{equation}
    \cM_e = 
    \left\{
    p(x;\theta) = 
    \exp 
    \left(
    \sum_{i=1}^{s} \theta_i t_i(x) - \psi(\theta)
    \right), \quad \theta = (\theta_1,\dots,\theta_s) \subseteq \bR^{s}
    \right\},
\end{equation}
which includes many important distributions such as the Gaussian distribution, exponential distribution, Poisson distribution, and Bernolli distribution, for example.

Let us consider a mixture family of distributions spanned by $d$ distinct probability functions $p_i(x)$,
\begin{align}
    \cM_m = 
    \left\{
    p(x;\theta) = \sum_{i=1}^{d} \theta_i p_i(x), \; \theta_i >0, \; \sum_{i=1}^{d} \theta_i = 1
    \right\}.
\end{align}
This set $\cM_m$ is closed under the internal division, i.e., any $m$-geodesic that connects two arbitrarily chosen distributions in $\cM_m$ is included in $\cM_m$. This means that the manifold is composed of straight lines and $\cM_m$ is a flat subset of $\cS$ in the sense of the straightness induced by $m$-geodesics. Similarly, for an exponential family, any $e$-geodesic connecting any two points in $\cM_e$ is included in $\cM_e$, and the subset $\cM_e$ is flat in the sense of the straightness induced by $e$-geodesics. The notion of flatness is defined in a more rigorous manner by using the metric and curvature tensors~\cite{amari2000methods,ay2017information}, but the above intuitive explanation suffices for explaining the $em$ algorithm in this paper. 

We then introduce the notion of orthogonal projection by defining tangent vectors and the inner product in the space of statistical model $\cS$. Consider the partial derivative operator $\partial_{\alpha} = \partial/\partial \alpha$ along with the direction $\alpha$, and as is conventionally done in the literature of differential geometry~\cite{kobayashi1996foundations}, we identify $\partial_{\alpha}$ as a basis of the tangent vector space for the manifold of interest. For example, a tangent vector along an $m$-geodesic with a parameter $t$ is
\begin{align}
    \partial_{t} \log r(x;t) =& 
    \frac{\partial_{t}r(x;t)}{r(x;t)} = \frac{q(x)-p(x)}{r(x;t)}.
\end{align}
A tangent vector along an $e$-geodesic is
\begin{align}
    \partial_t \log r(x;t) =& \log q(x) - \log p(x) - \dot{a}(t).
\end{align}
The tangent vectors of the model manifold are naturally defined by the derivatives with respect to the model parameter $\theta$ as
\begin{align}
    \partial_{i} \log p(x;\theta) = \frac{ \partial_i p(x;\theta)}{p(x;\theta)},
\end{align}
where $\partial_i$ is the partial derivative with respect to the $i$-th element of the parameter $\theta$. 
We can define a special form of the inner product in the space of probability distributions $\cS$ as
\begin{equation}
    \lan \partial_{\alpha} p, \partial_{\beta} p \ran 
    =
    \bE_{p} [ (\partial_{\alpha} \log p(X))
    (\partial_{\beta} \log p(X))
    ].
\end{equation}

Consider the point $p(\hat{\theta}) \in \cM$ closest to $q$ in terms of the KL divergence
\begin{align}
    \hat{\theta} = \argmin_{\theta}
    D(q , p(\theta)) =  \argmin_{\theta} \bE_{q}[\log q(X) - \log p(X;\theta)].
\end{align}
We assume that the model $p(x;\theta)$ is continuous with respect to both $x$ and $\theta$, and  partially differentiable with respect to the parameter $\theta$ in its domain; hence integrals and partial differentiations commute. Then, by definition, at $\theta = \hat{\theta}$, all partial derivatives of the KL divergence vanish:
\begin{align}
    \left. \partial_{i} D(q , p(\theta))\right|_{\theta=\hat{\theta}} = 
    - \bE_{q}
    \left[
    \partial_i \log p(X; \hat{\theta})
    \right] = 0.
\end{align}
The inner product of the tangent vector along the $m$-geodesic at $p(\hat{\theta})$ 
\begin{align}
\left.    \partial_t \log r(x;t) \right|_{t=0} =&
\left. 
\frac{q(x) - p(x;\hat{\theta})}{r(x;t)}
\right|_{t=0} \\
=&
\frac{q(x) - p(x;\hat{\theta})}{p(x;\hat{\theta})}
\end{align}
and the tangent vectors along the model manifold at $p(\hat{\theta})$ 
\begin{align}
\left.    \partial_{i} \log p(x;\theta) 
\right|_{\theta=\hat{\theta}} =
\frac{ \partial_i p(x;\hat{\theta})}{p(x;\hat{\theta})}
\end{align}
is calculated as
\begin{align}
    \bE_{p_{\hat{\theta}}}
    [
    \partial_t \log r(X;0) \cdot \partial_i \log p(X;\hat{\theta})
    ]
    =&
    \int 
    \left( 
    \frac{q(x) - p(x;\hat{\theta})}{p(x;\hat{\theta})}
    \right)
    \partial_i \log p(x;\hat{\theta}) p(x;\hat{\theta}) \dd x \\
    =& 
    \bE_q [ \partial_i \log p(X;\hat{\theta})]
    -
    \bE_{p_{\hat{\theta}}}
    [
    \partial_i \log p(X;\hat{\theta})
    ]=0.
\end{align}
Thus, the $m$-geodesic between $q$ and $p(\hat{\theta})$ is orthogonal to the model manifold, and $p(\hat{\theta})$ in this case is called the $m$-projection from $q$ onto $\cM$. 
We note that when $q$ is the empirical distribution $q(x) = \frac{1}{n} \sum_{i=1}^{n} \delta(x-x_i)$ of the observed data $\{x_i\}_{i=1}^{n}$, the $m$-projection coincides with the maximum likelihood estimation.

The $e$-projection is also defined in the same manner as the $m$-projection. 
Consider the point $p(\hat{\theta}) \in \cM$ closest to $q$ in terms of the KL divergence
\begin{align}
    \hat{\theta} = \argmin_{\theta}
    D(p(\theta) , q) = \bE_{p_{\theta}}[\log p(X;\theta) - \log q(X)].
\end{align}
By definition, 
\begin{align}
    \left. \partial_i D(p(\theta),q) \right|_{\theta=\hat{\theta}} 
    =&
    \int \partial_i p(x;\hat{\theta}) 
    (
    \log p(x;\hat{\theta}) - \log q(x)
    ) \dd x = 0.
\end{align}
The tangent vector along the $e$-geodesic is given by
\begin{align}
    \left. \partial_t  \log r(x;t) \right|_{t=0}
    =&
    \log q(x) - \log p(x;\hat{\theta}) - \left. \dot{a}(t)\right|_{t=0}\\
    =&
    \log q(x) - \log p(x;\hat{\theta}) -
    \bE_{p_{\hat{\theta}}}
    [
    \log q(X) - \log p(X;\hat{\theta})
    ].
\end{align}
Then, the inner product of this tangent vector and that of the model manifold is shown to be zero as
\begin{align}
    \bE_{p_{\hat{\theta}}}
    [
    \partial_t \log r(X;0) \cdot \partial_i \log p(X;\hat{\theta})
    ]
    =&
    \int \partial_i p(x;\hat{\theta}) 
    \left\{
    \log q(x) - \log p(x;\hat{\theta})
    \right\} \dd x=0,
\end{align}
so these two tangent vectors are orthogonal. 

It is known that the $e$-projection to an $m$-flat manifold is unique, and the $m$-projection to an $e$-flat manifold is also unique.

\begin{defi}
Let $\mathcal{M}$ be a submanifold of $\mathcal{S}$. Assume that for any $p,q \in \cM$ and $t \in (0,1)$, the element 
\begin{equation}
    r = t p + (1-t) q \in \cS
\end{equation}
belongs to $\cM$. Then, $\cM$ is said to be $m$-autoparallel. Let $\cE$ be a submanifold of $\cS$. Assume that for any $p,q \in \cE$ and $t \in (0,1)$, the element $r$ for which
\begin{equation}
    \log r = t \log p + (1-t) \log q - a(t)
\end{equation}
belongs to $\cE$, where the constant $a(t)$ is the normalizing factor. 
Then, $\cE$ is said to be $e$-autoparallel.
\end{defi}
We note that technically the notion of autoparallel is defined in terms of the covariant derivative~\cite{amari2000methods}, but the above definition suffices for the purpose of this paper.

\section{$em$ algorithm}
Consider the situation that a random vector $X$ is observed while there exists a hidden variable $Z$. The problem is to determine the parameter $\theta$ of the statistical model $p(x,z;\theta)$ only from the observations $\{x_1,\dots,x_n\}$. 
Since there are hidden variables that cannot be observed, it is impossible to calculate all the statistics needed to specify a point in the space $\cP$ only from the observed data. In this case, we first consider the marginal distribution of the observed variables and gather all the distributions that have the same marginal distribution as the empirical distribution of the observed variables. The set of these distributions conditioned by the marginal distributions represents observed data and is called the data manifold $\cD$. 
We introduce a parameter $\eta$ to specify the point in the data manifold $\cD$. Let $q(x)$ be the marginal distribution of $x$. All the points in $\cD$ have the same marginal distribution and any point in $\cD$ can be represented as 
\begin{align}
    q(x,z ; \eta) = q(x) q(z|x;\eta),
\end{align}
where $\eta$ is also regarded as the parameter of the conditional probability density function $q(z|x;\eta)$. 

A natural way of choosing a point in the model manifold $\cM$ is to adopt the closest point in $\cM$ from the data manifold $\cD$. It can be achieved by measuring the statistical distance between a point $q(\eta)$ in $\cD$ and a point $p(\theta)$ in $\cM$ with the KL divergence as
\begin{align}
    D(q(\eta), p(\theta)) = \int q(x,z;\eta) 
    \log \frac{
    q(x,z;\eta)
    }
    {
    p(x,z;\theta)
    }
    \dd x \dd z,
\end{align}
and obtaining the points $\hat{\eta}$ and $\hat{\theta}$ that minimize the divergence. The $em$ algorithm is a method of solving this estimation problem by applying the $e$-projection and the $m$-projection repeatedly. The procedure is composed of the following two steps.
\begin{description}
    \item[$e$-step:] Apply the $e$-projection from $\theta_t$ to $\cD$, and obtain $\eta_{t+1}$ by
    \begin{align}
        \eta_{t+1} = \argmin_{\eta} D(q(\eta), p(\theta_t)).
        \label{eq:e_step}
    \end{align}
    \item[$m$-step:] Apply the $m$-projection from $\eta_{t+1}$ to $\cM$ and obtain $\theta_{t+1}$ by
    \begin{align}
        \theta_{t+1} = \argmin_{\theta} D(q(\eta_{t+1}), p(\theta)).
        \label{eq:m_step}
    \end{align}
\end{description}
Starting from an initial value $\theta_0$, the procedure is expected to converge to the optimal value after a sufficiently large number of iterations.

\begin{figure}[ht]
\centering
  \includegraphics[width=.8\linewidth]{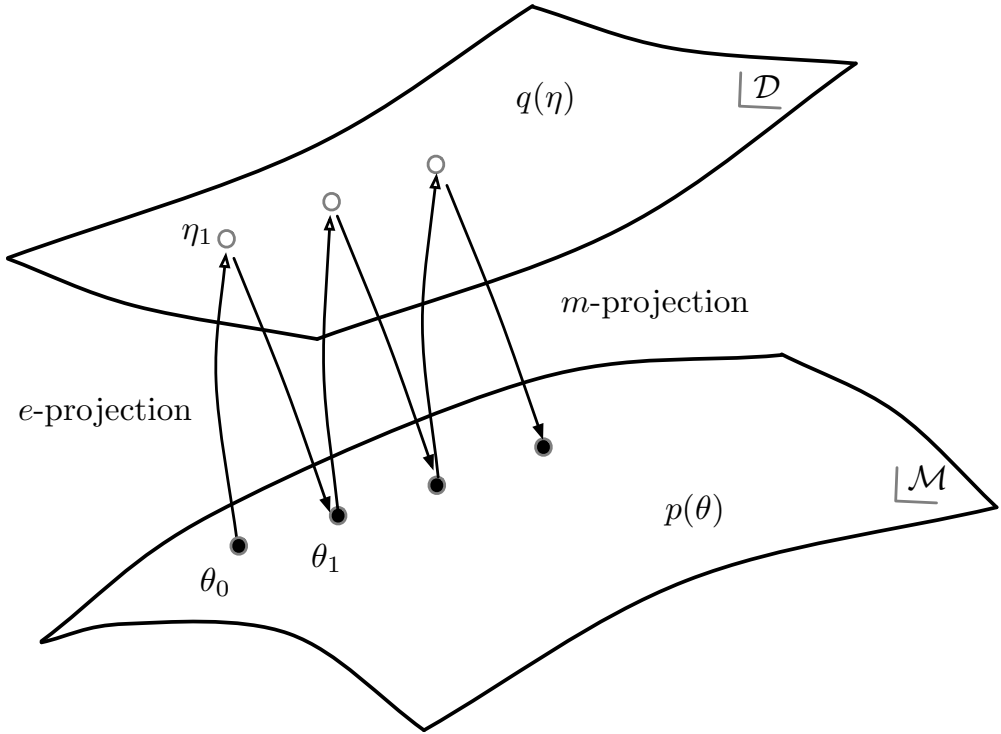}
  \caption{Geometric perspective of the $em$ algorithm.}
  \label{fig:em}
\end{figure}

If the model manifold is $e$-flat and the data manifold is $m$-flat, it is shown that in each step, the projection is uniquely determined, but the algorithm can converge to one of the local minima in general.

Note that the procedure in the $e$-step is equivalent to minimizing 
\begin{align}
    D(q(\eta), p(\theta)) =&
    \int q(x) q(z|x;\eta) 
    \log 
    \frac{
    q(x) q(z|x;\eta)
    }{
    p(x;\theta_t) p(z|x;\theta_t)
    }
    \dd x \dd z \\
    =&
    \int q(x) \log \frac{q(x)}{p(x;\theta_t)} \dd x \\
    &+ \int 
    q(x) q(z|x;\eta) \log
    \frac{
    q(z|x;\eta)
    }{
    p(z|x;\theta_t)
    }
    \dd x \dd z\\
    =&
    \int q(x) \log \frac{q(x)}{p(x;\theta_t)} \dd x \\
    &+ \int 
    q(x) D(q(z|x;\eta), p(z|x;\theta_t)) \dd x.
\end{align}
It is reduced to minimizing the conditioned KL divergence $D(q(z|x;\eta), p(z|x;\theta_t))$. Because of the positivity of the KL divergence, in most cases, the parameter update $\eta_{t} \to \eta_{t+1}$ is realized by solving
\begin{equation}
    q(z|x;\eta_{t+1}) = p(z|x;\theta_{t})
\end{equation}
with respect to $\eta_{t+1}$. 

Remember that the EM algorithm is an alternating optimization procedure composed of the E and M steps.
\begin{description}
    \item[E-step:] Calculate $Q(\theta,\theta_t)$ defined by
    \begin{align}
        Q(\theta,\theta_t) = 
        \frac{1}{n} \sum_{i=1}^{n} 
        \left\{
        \int p(z|x_i; \theta_t) \log p(x_i,z; \theta) \dd z
        \right\}.
    \end{align}
    \item[M-step:] Find $\theta_{t+1}$ that maximizes $Q(\theta,\theta_t)$ with respect to $\theta$:
    \begin{equation}
        \theta_{t+1} = \argmax_{\theta} Q(\theta,\theta_t).
    \end{equation}
\end{description}
The EM algorithm can be also seen as a motion on the data manifold and the model manifold. In the M-step, the estimate is obtained by the $m$-projection from a point in the data manifold to a point in the model manifold, and this operation is equivalent to the $m$-step. On the other hand, in the E-step, the conditional expectation is considered and this is slightly different from the $e$-projection in the $e$-step. 

Let $q(x)$ be the empirical distribution of the observed variables $X$. Suppose $q(z|x;\eta_{t+1}) = p(z|x;\theta_{t})$ holds in the $e$-step, then the objective function evaluated in the $m$-step is
\begin{align}
    D(q(\eta_{t+1}), p(\theta)) =&
    \int q(x)p(z|x;\theta_t)
    \log 
    \frac{
    q(x) p(z|x;\theta_t)
    }{
    p(x,z;\theta)
    }
    \dd x \dd z\\
    =&
    \int q(x) p(z|x;\theta_t) 
    \log q(x) p(z|x;\theta_t) \dd x \dd z - Q(\theta,\theta_t).
\end{align}
This shows that the $m$-step and the $M$-step are equivalent if the first term can be properly integrated. The problem occurs when the integrals including the empirical distribution, which is sum of delta functions, are not appropriately defined. In~\cite{AMARI19951379}, the case where $\cP$ is an exponential family and the model manifold is a curved exponential family embedded in $\cP$ was considered, and it was shown that the E-step and the $e$-step give different estimates. This result mainly comes from the fact that the expectation of the hidden variables and the expectation conditioned by the observed variables do not agree:
\begin{equation}
    \bE_{q(\eta)}[Z] \neq \bE_{q(\eta)}[ Z|x=\bE_{q(\eta)}[X]].
\end{equation}

\subsection{Robust variant: $um$ algorithm}
Since the EM algorithm is an algorithm for maximum likelihood estimation, in this paper, we mainly consider KL divergence. However, it is well known that KL divergence is vulnerable to outliers, as is maximum likelihood estimation, and robust estimation methods have been proposed using the Bregman divergence~\cite{BREGMAN1967200}.

Let $U$ be a monotonically increasing convex function on $\bR$, and $u$ be the derivative of $U$. We define $U^{\ast}(\zeta) = \sup_{z \in \mathbb{R}} \{z \zeta -U(z)\}$, that is, the Legendre transform of $U$, and $u^{\ast} = u^{-1}$ as the derivative of $U^{\ast}$. We consider transforming the function $f$ by $u^{\ast}(f)$ and denote it as $\breve{f} = u^{\ast}(f)$, which is called the $u$-representation of the function $f$. Then, the Bregman potential between two functions $f$ and $g$ is defined as
\begin{align}
    d_U(f,g) = U^{\ast}(f) + U(\breve{g}) - f \breve{g},
\end{align}
and the Bregman divergence is defined as
\begin{align}
    D_{U}(p,q) = 
    \int d_{U}(p(y),q(y)) \dd \Lambda (y) = \int d_{U}(p,q) \dd \Lambda,
\end{align}
where $p$ and $q$ are probability density or probability mass functions. 
Note that we omit the integral variable $y$ for notational simplicity. 

The most popular convex function $U$ and its related functions for Bregman divergence would be the exponential function, which leads to the KL divergence where
\begin{align}
\begin{aligned}
    U(z) &= \exp(z), 
    & U^{\ast}(\zeta) &= 
    \zeta ( \log \zeta -1 ),
\\
    u(z) &= \exp(z),&
    u^{\ast}(\zeta) &= \log \zeta.
\end{aligned}
\end{align}

Other important examples include the $\eta$-type with $\eta \geq 0$
\begin{align}
\begin{aligned}
    U(z) &= \exp(z) + \eta z, &
    U^{\ast}(\zeta) &= (\zeta - \eta) \{ \log (\zeta - \eta) +1\},\\
    u(z) &= \exp(z) + \eta, &
    u^{\ast}(\zeta) &= \log (\zeta - \eta),
\end{aligned}
\end{align}
and the $\beta$-type with $\beta \geq 0$
\begin{align}
\begin{aligned}
    U(z) &= \frac{1}{\beta+1} (\beta z + 1)^{\frac{\beta+1}{\beta}}, 
    & U^{\ast}(\zeta) &= 
    \frac{\zeta^{\beta+1}}{\beta(\beta+1)} - \frac{\zeta}{\beta},\\
    u(z) &= (\beta z + 1)^{1/\beta},&
    u^{\ast}(\zeta) &= \frac{\zeta^{\beta} -1}{\beta}.
\end{aligned}
\end{align}
Both the $\eta$-type and $\beta$-type functions are known to lead to robust estimators. 

The Bregman divergence is also called the $u$-divergence, and the robust variant of the $em$ algorithm based on the Bregman divergence is called the $um$ algorithm. 
The basic idea is simply to change the $e$-projection to $u$-projection, i.e., instead of Eq.~\eqref{eq:e_step} in the $em$ algorithm, we consider 
\begin{equation}
    \psi^{(t+1)} = \argmin_{\psi} D_{U}(p(\psi), q(\theta^{(t)})).
\end{equation}
However, $u$-projections with respect to Bregman divergences such as $\beta$-divergence and $\eta$-divergence are generally not obtained in closed form. In~\cite{FujimotoMurataEM2007}, for estimating the model and mixture parameters in finite mixture models, two simplifications of the $m$ projection were proposed to make the inference computationally feasible.
The influence function of the $u$-mixture of the exponential family models with respect to the outlying mixture component was derived in~\cite{HE_IG_2022}. We also note that the extension to the Bregman divergence is reconsidered in~\cite{DBLP:journals/corr/abs-2201-02447} and applied to the rate distortion problem in the quantum channel.

\section{Geometric perspective of channel capacity}
In this section, we introduce the information geometric perspective of the estimation algorithm of channel capacity.

A memoryless channel with finite input alphabet $\Omega_1$ and finite output alphabet $\Omega_2$ is determined by a stochastic matrix $R : \Omega_1 \to \Omega_2$ or a family of distributions $\{r(\cdot | x )\}_{x \in \Omega_1}$ on $\Omega_2$.

Let $\cS_i$ be the sets of all probability distributions on $\Omega_i, i=1,2$:
\begin{align}
    \cS_i = \{ p : \Omega_i \to \bR_{++} \mid 
    \sum_{x \in \Omega_i} p(x) = 1\}, \; i=1,2,
\end{align}
where $\bR_{++} = \{ x \in \bR \mid x >0\}$. Similarly, let $\cS_3$  be the set consisting of all probability distributions on $\Omega_1 \times \Omega_2$.
A channel is defined by a triple $(\Omega_1, r(y|x), \Omega_2)$ of finite sets $\Omega_1, \Omega_2$ and a map $r : x \mapsto r(\cdot|x)$. The map $I:\cS_3 \to \bR$ defined by
\begin{equation}
    I(p(x,y)) = D(p(x,y) , q(x) \cdot r(y))
\end{equation}
is called the mutual information, where $q(x)$ and $r(y)$ are marginal distributions of $p(x,y) \in \cS_3$. Given a channel $(\Omega_1, r(y|x), \Omega_2)$, the channel capacity is defined by 
\begin{align}
    C = \sup_{q(x) \in \cS_{1}} I(q(x) \cdot r(y|x)).
\end{align}

Suppose that a probability distribution $\hat{q}(x) \in \cS_1$ attains the channel capacity $C$. Then for any $x \in \Omega_1$, the following equation holds:
\begin{align}
    D(r(y|x), r_{\hat{q}}(y)) = C
    \label{eq:CC_attain1}
\end{align}
where $r_{\hat{q}}(y)$ is the marginal distribution of $\hat{q}(x) \cdot r(y|x)$ on $\Omega_2$. Conversely, if there exist $\hat{C} \geq 0$ and $\hat{q} \in \cS_1$ satisfying
\begin{align}
    D(r(y|x), r_{\hat{q}}(y)) = \hat{C}
    \label{eq:CC_attain2}
\end{align}
for all $x \in \Omega_1$, then $\hat{C} \geq 0$ and $\hat{q}(x)$ are the channel capacity and a probability distribution that attains the channel capacity, respectively.

The Arimoto algorithm~\cite{1054753} updates a distribution $q^{(t)}(x) \in \cS_1$ by the update rule
\begin{align}
    q^{(t+1)}(x) =
    \frac{
    q^{(t)}(x) \exp \{ D(r(y|x) , r^{(t)}(y))\}
}{
\sum_{x'} q^{(t)} (x') \exp \{ D(r(y|x') , r^{(t)}(y))\}
},
\label{eq:Arimoto}
\end{align}
where $r^{(t)}(y)$ is the marginal distribution of $q^{(t)} (x) \cdot r(y|x)$ and is denoted as $r^{(t)}(y)= r_{q^{(t)}}(y)$. 
The channel capacity $C$ of a discrete memoryless channel is shown to be concave and $\cS_1$ is a convex set. It is proven that the Arimoto algorithm monotonically increases the mutual information $I(q^{(t)}(x) \cdot r(y|x))$, which converges to the channel capacity~\cite{yeung2008information}.

\subsection{Information geometric perspective of channel capacity}
Recently, the information geometric perspective of the Arimoto algorithm has been elucidated~\cite{Toyota2020}. 

Define subsets $\cM$ and $\cE$ of $\cS_3$ as
\begin{align}
    \cM =& \{ q(x) \cdot r(y|x) \mid q(x) \in \cS_1\},\\
    \cE =& \{ \tilde{q}(x) \cdot r(y) \mid \tilde{q}(x) \in \cS_1, r(y) \in \cS_2\}.
\end{align}
The subspace $\cM$ is composed of probability distributions of the form $q(x) \cdot r(y|x)$. The conditional distribution $r(y|x)$ is a fixed channel; hence any point in $\cM$ is specified by an input distribution $q(x) \in \cS_1$. In contrast, $\cE$ is composed of probability distributions of the form $\tilde{q}(x) \cdot r(y)$, namely, the distributions of the input and the output are mutually independent. Note that any input distribution in $\cE$ is denoted by $\tilde{q}(x) \in \cS_1$ to differentiate it from that in $\cM$. 

It is easy to verify that $\cM$ is $m$-autoparallel and $\cE$ is $e$-autoparallel. For $p(x,y) \in \cS_3$, the $m$-projection of $p$ onto $\cE$ is $q(x)\cdot r(y)$. 
Then, the capacity is written as
\begin{align}
    C = \sup_{p(x,y) \in \cM} 
    D
    (p(x,y) ,
    \Pi^{(m)} (p(x,y))),
\end{align}
where $\Pi^{(m)}(p(x,y))$ is the $m$-projection of $p(x,y)$ onto $\cE$. From this expression, we see that the channel capacity $C$ is characterized by the largest divergence from $\cM$ to $\cE$. 

In contrast to the EM algorithm, for estimating the channel capacity, we must maximize the KL divergence between two flat statistical manifolds. 
We cannot expect convergence to the channel capacity by a simple iteration of $e$ and $m$ projections in the $em$ algorithm. For this problem, the inverse $em$ algorithm was proposed in~\cite{Toyota2020}.

For $q^{(t)}(x) \cdot r(y|x) = p^{(t)} (x,y) \in \cM$, update $q^{(t+1)}(x) \cdot r(y|x) = p^{(t+1)}(x,y) \in \cM$ as follows.
\begin{description}
    \item[Backward $e$-step:] Search $\tilde{q}^{(t+1)}(x) \cdot r^{(t+1)}(y) \in \cE$ such that the unique $e$-projection from $\tilde{q}^{(t+1)}(x) \cdot r^{(t+1)}(y)$ onto $\cM$ is $p^{(t)}(x,y)$.
    \item[Backward $m$-step:] Search $q^{(t+1)}(x) \cdot r(y|x) \in \cM$ such that the unique $m$-projection from $q^{(t+1)}(x) \cdot r(y|x)$ onto $\cE$ is $\tilde{q}^{(t+1)}(x) \cdot r^{(t+1)}(y)$. Set $p^{(t+1)}(x,y) = q^{(t+1)}(x)\cdot r(y|x)$.
\end{description}

\begin{figure}
\centering
  \includegraphics[width=.8\linewidth]{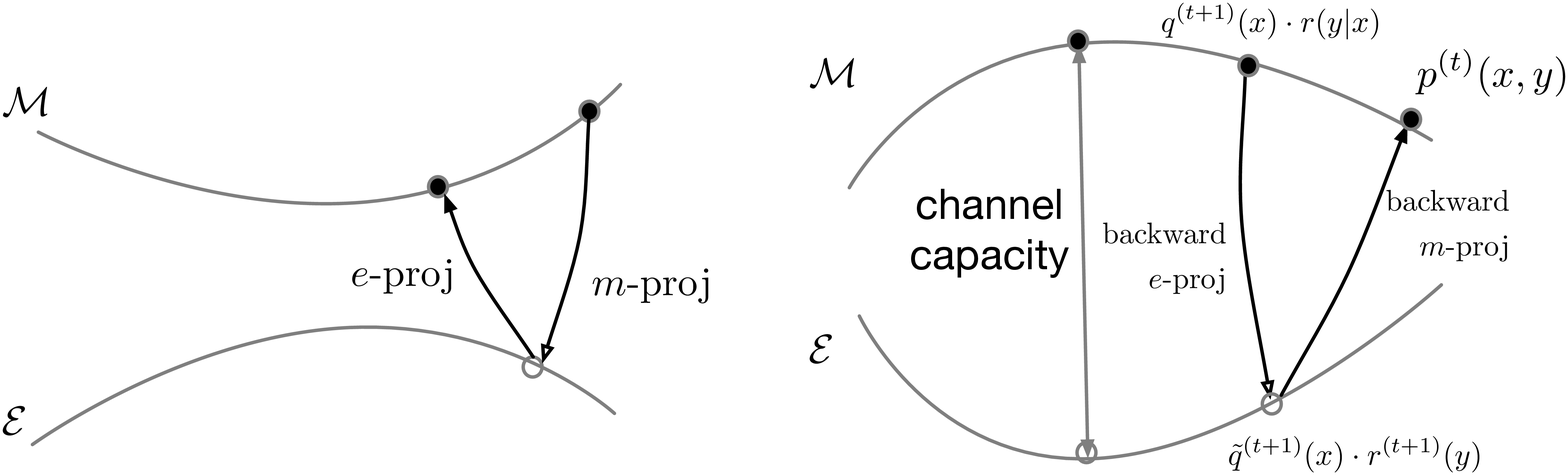}
  \caption{Left: The $em$ algorithm to minimize the KL divergence between two manifolds. Right: The backward $em$ algorithm to maximize the KL divergence between two manifolds.}
  \label{fig:Bem}
\end{figure}

It is proven that 
\begin{equation}
    I(p^{(t)}(x,y)) \leq I(p^{(t+1)}(x,y))
\end{equation}
holds. 

For the backward $e$-step, it is shown that for a given probability distribution $p^{(t)}(x,y) \in \cM$, there exists a probability distribution $\tilde{q}^{(t+1)}(x) \cdot r^{(t+1)}(y) \in \cE$ that satisfies $\Pi^{(e)} (\tilde{q}^{(t+1)}(x) \cdot r^{(t+1)}(y)) = p^{(t)}(x,y)$, and it is written as
\begin{align}
    \tilde{q}^{(t+1)}(x) \propto q^{(t)}(x) \exp 
    \{
    D(r(\cdot|x) , r(\cdot))
    \}.
    \label{eq:q_in_E}
\end{align}
For later use, define an $e$-autoparallel subset $\cE^{(t)}$ of $\cE$ by
\begin{align}
    \cE^{(t)} = 
    \{
    \tilde{q}(x) \cdot r(y) \mid 
    \Pi^{(e)} ( \tilde{q}(x) \cdot r(y)) = q^{(t)} (x) \cdot r(y|x) \in \cM
    \},
\end{align}
which is composed of candidates for the backward $e$-step. 

\begin{figure}
\centering
  \includegraphics[width=.8\linewidth]{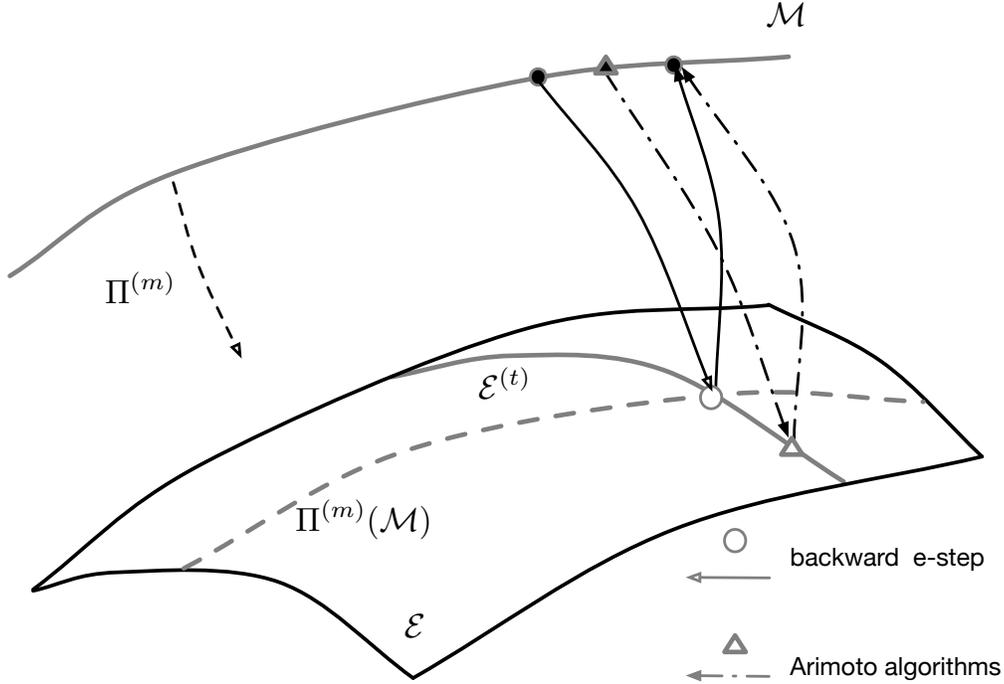}
  \caption{Schematics of backward $em$ and Arimoto algorithms. In the backward $e$-step, the intersection of $\Pi^{(m)}(\cM)$ and $\cE^{(t)}$ is searched. In contrast, the Arimoto algorithm only considers the restriction $\cE^{(t)}$ for update.}
  \label{fig:Arimoto}
\end{figure}

To carry out the backward $m$-step, it is important to choose an appropriate probability distribution $\tilde{q}^{(t+1)}(x) \cdot r^{(t+1)}(y) \in \cE^{(t)}$ so that there exists $p^{(t+1)}(x,y) \in \cM$ such that $\Pi^{(m)}(p^{(t+1)}(x,y)) = \tilde{q}^{(t+1)}(x) \cdot r^{(t+1)}(y)$. Let $\Pi^{(m)}(\cM)$ be the projection of $\cM$ to $\cE$ by the $m$-projection (Fig.~\ref{fig:Arimoto}), and assume\footnote{The existence and uniqueness of the intersection are not guaranteed in general.} there exist intersections of $\Pi^{(m)}(\cM)$ and $\cE^{(t)}$. Choose an arbitrary point $\tilde{q}^{(t+1)}(x) \cdot r^{(t+1)}(y) \in \Pi^{(m)}(\cM) \cap \cE^{(t)}$. For such a point, we can perform the backward $m$-step.

The problem of finding $\tilde{q}^{(t+1)}(x) \cdot r^{(t+1)}(y) \in \cE^{(t)}$ for the backward $m$-step is equivalent to finding an intersection of $\Pi^{(m)}(\cM) \cap \cE^{(t)}$. Let us focus on element $\tilde{q}^{(t+1)}(x) \cdot r^{(t+1)}(y) \in \cE^{(t)}$. Given $r^{(t+1)}(y)$, the form of $\tilde{q}^{(t+1)}(x)$ is determined by Eq.~\eqref{eq:q_in_E}; hence it only depends on $r^{(t+1)}(y)$, and $\tilde{q}^{(t+1)}(x)$ is regarded as a function of $r^{(t+1)}(y)$ henceforth. 
Note that by the definition of $m$-projection to $\cE$, $\Pi^{(m)}(q^{(t+1)}(x) \cdot r(y|x)) = \tilde{q}^{(t+1)}(x) \cdot r_{q^{(t+1)}}(y)$, where $r_{q^{(t+1)}}(y)$ is the marginal distribution of $q^{(t+1)}(x)\cdot r(y|x)$.
Then, the function $r^{(t+1)}(y)$ must satisfy the following condition.
\begin{align}
    \exists \tilde{q}^{(t+1)}(x) \in \cS_1 \; s.t. \;
    \Pi^{(m)}(q^{(t+1)}(x) \cdot r(y|x)) = 
    \tilde{q}^{(t+1)}(x) \cdot r_{q^{(t+1)}}(y) = 
    q^{(t+1)}(x) \cdot r^{(t+1)}(y).
\end{align}
Concretely, 
\begin{align}
    r_{q^{(t+1)}}(y) = &
    \sum_{x \in \Omega_1} 
    q^{(t+1)}(x) \cdot r(y|x) \\
    \propto &
    \sum_{x \in \Omega_1} 
     q^{(t)}(x)
    \cdot 
    \exp D(r(\cdot | x), r^{(t+1)}(\cdot))
    \cdot r(y|x), 
\end{align}
hence we must solve 
\begin{align}
r^{(t+1)}(y) =
    \frac{1}{Z(r^{(t+1)})}
      \sum_{x \in \Omega_1} 
       \tilde{q}^{(t)}(x)
    \cdot 
    \exp D(r(\cdot | x), r^{(t+1)}(\cdot))
    \cdot r(y|x)
    \label{eq:BeStep}
\end{align}
with respect to $r^{(t+1)}(y)$, where 
$Z(r^{(t+1)})$ is the normalization term. This problem of finding the distribution $r^{(t+1)}(y)$ by solving Eq.~\eqref{eq:BeStep} is prohibitive in general. To make the problem tractable, consider approximating the KL divergence in Eq.~\eqref{eq:BeStep} by a constant. It is, by using Eq.~\eqref{eq:CC_attain1} and~\eqref{eq:CC_attain2}, regarded as approximating $D(r(\cdot |x), r^{(t+1)}(\cdot))$ by the attained channel capacity $C$. Then, the problem in Eq.~\eqref{eq:BeStep} is reduced to
\begin{align}
    r^{(t+1)}(y) = \sum_{x \in \Omega_1} q^{(t)}(x) \cdot r(y|x) = r_{q^{(t)}}(y),
\end{align}
which is the explicit solution of $r^{(t+1)}(y)$, and $\tilde{q}^{(t+1)}(x) \in \cS_1$ is also approximated as 
\begin{equation}
    \tilde{q}^{(t+1)}(x) \appropto q^{(t)}(x) \exp D(r(\cdot |x), r_{q^{(t)}}(\cdot)).
\end{equation}
The backward $m$-step also must be approximated because, owing to the approximation of the backward $e$-step, $\tilde{q}^{(t+1)}(x) \cdot r_{q^{(t)}}(y)$ is not necessarily in $\Pi^{(m)}(\cM) \cap \cE^{(t)}$. The backward $m$-step is simply approximated by the $m$-projection of  $\tilde{q}^{(t+1)}(x) \cdot r_{q^{(t)}}(y)$ to $\cM$ and given as
\begin{align}
\Pi^{(m)}(\tilde{q}^{(t+1)}(x) \cdot r_{q^{(t)}}(y)) = \tilde{q}^{(t+1)}(x) \cdot r(y|x).
\end{align}
In summary, the approximated backward $em$ algorithm is reduced to the updates of $\tilde{q}^{(t+1)}(x)$ by
\begin{equation}
    \tilde{q}^{(t+1)}(x) \propto
    q^{(t)}(x) \exp D(r(\cdot |x), r_{q^{(t)}}(\cdot)),
\end{equation}
which is nothing but the Arimoto algorithm~\eqref{eq:Arimoto}.

\subsection{Addendum: turbo decoding, LDPC code}
Finally, we mention the information geometric approach for other instances of information theory. Turbo codes and low-density parity check (LDPC) codes have revolutionized code theory research and are now in practical use and standardized. The common features of these codes are that they are composed of simple codes and that they can be decoded with low computational complexity even when the code length is large. In addition, by designing appropriately long codes, it is possible to achieve a channel capacity close to the theoretical bound. 
Turbo decoding is a method of maximum posterior marginal decoding of codes passing a memoryless binary symmetric channel using two parity check words. It has a special iterative estimation structure. This iterative structure is different from that of the EM algorithm, but it is also analyzed precisely from the viewpoint of information geometry~\cite{IkedaTA2004}.

\section{Parameter estimation of statistical models with structures}
Various structures in the data distribution space can be modeled flexibly and naturally using statistical models. As an example, we introduce the problem of item preference parameter estimation, in which parameters on a probability simplex representing the ordinal structure of a finite number of items are estimated from observations on item pairs, and show that the problem can be solved using the $em$ algorithm.

The mode of probability distribution is useful as a location parameter to characterize the distribution structure, but it is more difficult to handle than the expectation. In the latter half of this section, we introduce modal linear regression, a linear regression on the mode, and geometrically construct the $em$ algorithm for estimating the regression coefficients.

The Boltzmann machine with hidden layers is a popular neural network generative model. The parameter estimation problem of Boltzmann machines is also formulated as the minimization of the KL divergence between two statistical manifolds, and its geometric structure is studied.

\subsection{Preference parameter estimation in ranking models}
Given a set of rating data for a set of items $\{I_1,\dots,I_N\}$, determining preference levels of items is an important problem. Various probability models for preference have been proposed. As an example, in~\cite{r.a.bradley52:_rank_analy_of_incom_block_desig}, the Bradley--Terry (BT) model was proposed, in which each item $I_i$ has a positive-valued parameter $\theta_i$, and the probability of being chosen item $I_i$ over item $I_j$, which is denoted by $I_i \succ I_j$, is given by $\Pr(I_i \succ I_j) = \frac{\theta_i}{\theta_i+\theta_j}$. 
Namely, we consider a parameter set $Q = \{\theta_i \}_{i \in \Lambda}, \; \sum_{i \in \Lambda} \theta_i = 1, \; \theta_i >0$, where $\Lambda =\{1,2,\dots,N\}$ is an index set. 
In this model, the greater the value of $\theta_i$, the more highly item $I_i$ is preferred. 
Assume that multiple users independently compare $i$ and $j$, and let $n_{ij}$ and $n_{ji}$ be the number of observed events $I_i \succ I_{j}$ and $I_j \succ I_i$, respectively. The log-likelihood of the BT model is given by
\begin{equation}
    L(Q) = \sum_{i \neq j} n_{ij} \log \frac{\theta_i}{\theta_i + \theta_j},
\end{equation}
and the estimate $\tilde{Q}$ is obtained as a solution of the following optimization problem:
\begin{equation}
    \tilde{Q} = \argmax_{Q} L(Q) \quad \text{subject to} \quad \sum_{i \in \Lambda} \theta_i =1, \; \theta_i >0.
\end{equation}
There exists several parameter estimation algorithms~\cite{10.1214/aos/1028144844,NIPS2004_825f9cd5}.

\begin{figure}[ht]
\centering
  \includegraphics[width=.9\linewidth]{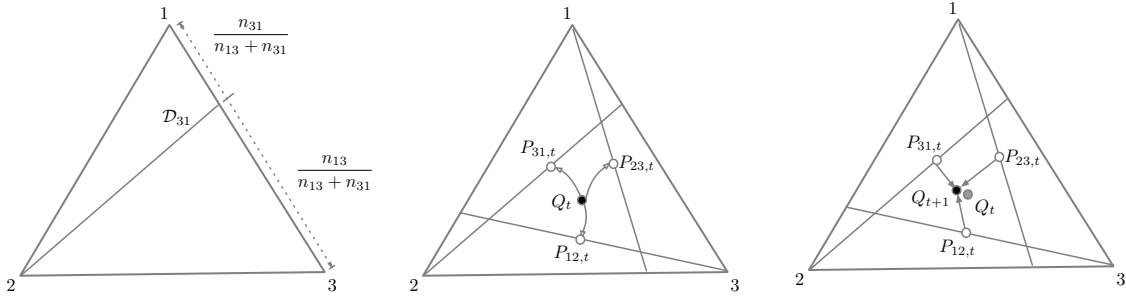}
  \caption{Left: Example of the $m$-flat data manifold embedded in the two-probability simplex. Middle: $e$-projection from $Q_t$ in $\cM$ to data manifolds $\cD_{ij}$. Right: $m$-projection from data manifolds to the model manifold.}
  \label{fig:BT}
\end{figure}

We can take another look at the BT estimation problem from the viewpoint of information geometry. Consider a space of categorical distributions 
\begin{align}
    \cM = \left\{ 
    Q=\{\theta_i\}_{i \in \Lambda} 
    \; \middle|  \;
    \sum_{i\in \Lambda} \theta_i = 1, \; \theta_i >0
    \right\}.
\end{align}
Consider also a set of probabilities $P = \{ \pi_i\}_{i \in \Lambda}$. Items $I_i$ and $I_j$ are compared several times, and we observe the event $I_i \succ I_j$  $n_{ij}$ times and the event $I_j \succ I_i$ $n_{ji}$ times. This observation $(n_{ij},n_{ji})$ indicates a restriction for probabilities in the BT model as $\pi_i : \pi_j = n_{ij} : n_{ji}$. For the observation $(n_{ij}, n_{ji})$, we define a subspace $\cD_{ij}$ of $\cM$ that satisfies the observed ratio as 
\begin{align}
    \cD_{ij} = \{ 
    P = \{ \pi_{i} \}_{i \in \Lambda} \in \cM 
    \mid 
    \pi_i : \pi_j = n_{ij} : n_{ji}
    \}.
\end{align}
This submanifold $\cD_{ij}$ gives a constraint on the simplex in accordance with the observation $(n_{ij},n_{ji})$, as shown in Fig.~\ref{fig:BT} (left panel), and is called the data manifold. The data manifold $\cD_{31}$ is, for example, composed of the count $(n_{13}, n_{31})$ of the events $I_1 \succ I_3$ and  $I_3 \succ I_1$. It divides the edge of items $I_1$ and $I_3$ to the ratio $n_{31}:n_{13}$. 

If the set of data manifolds $\{\cD_{ij}\}$ correspond to all observations of the form $(n_{ij}, n_{ji}), \; i,j \in \Lambda$ is consistent, that is, it has a unique intersection, and it is adopted to be the estimate $\tilde{Q} = \cap_{i,j} D_{ij}$. However, this is not the case in general, and it is reasonable to seek a model that is maximally consistent with the observed pairwise comparison data. Let $N$ be the number of given data manifolds $\{ \cD_{ij}\}$. A good estimate for the BT models is obtained as the nearest point in the simplex from these $N$ submanifolds. A natural choice of the measure of closeness in the simplex is the KL divergence. The KL divergence between points $P=\{\pi_i\}$ and $Q=\{\theta_i\}$ is given as
\begin{equation}
    D(P,Q) = \sum_{i \in \Lambda} \pi_i \log \frac{\pi_i}{\theta_i}.
\end{equation}
On the basis of this, we define the KL divergence between a submanifold $\cD$ and a point $Q$ as
\begin{equation}
    D(\cD,Q) = \min_{P \in \cD} D(P,Q).
\end{equation}
Then, an objective function for the parameter estimation on the simplex $\cM$ is proposed as the average of the KL divergences between $\cD_{ij}$ and $Q$ as
\begin{align}
    F(Q) =\frac{1}{N} \sum_{i,j} D(\cD_{ij},Q) = 
    \frac{1}{N} \sum_{i,j} \min_{P \in \cD_{ij}} D(P,Q),
\end{align}
and the minimizer of this function $F(Q)$ is obtained by solving the following optimization problem:
\begin{align}
    \hat{Q} = 
    \argmin_{Q \in \cM} 
    \left\{
    \sum_{i,j} \min_{P \in \cD_{ij}} D(P,Q)
    \right\}.
\end{align}
This is a nested optimization problem and direct optimization is difficult, and the $em$ algorithm is applicable to solve this problem. 

The data manifold $\cD_{ij}$ is defined as the ratio of the observed pairwise comparisons $n_{ij}$ and $n_{ji}$, hence it is an $m$-flat manifold. In~\cite{DBLP:journals/neco/FujimotoHM11}, it is shown that there exists an $e$-flat subspace $\cS (P)$ in $\cM$ for an arbitrary point $P \in \cD_{ij}$ and, conversely, an arbitrary point $Q \in \cM$ has a unique point $P \in \cD_{ij}$ such that $Q \in \cS (P)$ holds. Based on these flat structures, it is guaranteed that the $e$-projection from $\cD_{ij}$ to $Q \in \cM$ defined as
\begin{equation}
    \hat{P}_{ij} = \argmin_{P \in \cD_{ij}} D(P,Q)
\end{equation}
and the $m$-projection from $\cM$ to a set of points $\{ P_{ij} \in \cD_{ij}\}$ defined as
\begin{equation}
    \hat{Q} = \argmin_{Q \in \cM} \sum_{i,j} D(\hat{P}_{ij}, Q)
\end{equation}
are uniquely determined.

In summary, the $em$ algorithm for estimating the preference parameter of the BT model is given as follows. Starting from an initial parameter $Q_0$, set $t=0$, and repeat the $e$- and the $m$-steps.
\begin{description}
    \item{$e$-step:}
    For each $(i,j)$, find a point in $\cD_{ij}$ by the $e$-projection
    \begin{align}
        \hat{P}_{ij,t} = \argmin_{P \in \cD_{ij}} D(P,Q_t).
    \end{align}
    \item{$m$-step:}    
    Find a point $Q_{t+1}$ that is the closest to $\cD_{ij}$ by $m$-projection:
    \begin{align}
        Q_{t+1} = \argmin_{Q \in \cM} \sum_{i,j}
        D(P_{ij,t}, Q).
    \end{align}
\end{description}
The $e$-projection is depicted in Fig.~\ref{fig:BT} (Middle) and the $m$-projection is depicted in Fig.~\ref{fig:BT} (Right).

The natural extension of the BT model to the multiple comparison is given by Plackett~\cite{Plackett1975} and we refer to the model as the Plackett--Luce model,
\begin{equation}
    \Pr(I_{a(1)} \succ I_{a(2)} \succ \cdots \succ I_{a(N)}) = 
    \prod_{i=1}^{N-1} 
    \frac{
    \theta_{a(i)}
    }{
    \sum_{j=i}^{N} \theta_{a(j)}
    },
\end{equation}
where $a(j)$ denotes the index of the item that occupies the $j$-th position in the ranking, and its geometric properties are also investigated. It is further generalized~\cite{DBLP:conf/pakdd/HinoFM09,DBLP:journals/neco/HinoFM10} to cope with the grouped ranking observation, in which each of $U$ judges rates $N$ items on a scale of $1$ to $M$, $M\leq N$, assuming there is a latent ordering in a set of the same rated items, but we only observe $M$ groups of items that are divisions of $N$ items. The grouping by a user $u$ is denoted as $D_u = \{G_1^u,\dots,G_M^u\}$ where $G^u_m = \{ i \in \{1,\dots,N\} | I_i \in m\mbox{-th group}\}$. The problem of finding the optimal parameter $\theta$ most consistent with the observations $\{D_{u}\}_{u=1}^{U}$ is, also in this case, solved using the $em$ algorithm. 

Explaining preference levels of all users by a single set of preference parameter is not reasonable. In~\cite{DBLP:journals/neco/HinoFM10}, a mixture of different preference parameters and user clustering based on the mixture model was proposed. 
The application of the mixture model to the visualization of the item--user relationship and item recommendation were also proposed~\cite{DBLP:conf/ic3k/FujimotoHM09,DBLP:journals/neco/HinoFM10}. 
Moreover, in~\cite{DBLP:journals/neco/FujimotoHM11}, a weight for an observation was introduced to reflect the reliability of each observation, and the sensitivity to the outlying observation was also analyzed. 

\subsection{Modal linear regression model}

Linear regression is used to model the conditional mean of a response variable $y$ given the predictor variable $x$. A well-known least-squares estimator for linear regression coefficients is highly sensitive to outliers. 
To alleviate this problem, many estimators have been developed. 
One of the reasons for this sensitivity stems from the mean estimation. Mode is a reasonable alternative to characterize the location of distributions and is used to, for example, robustly identify low-dimensional subspace~\cite{DBLP:journals/neco/SandoH20}. 
Modal linear regression~(MLR;\cite{Lee1989}) is used to model the conditional mode of $y$ given $x$ by using a linear predictor function of $x$. MLR relaxes the distribution assumptions for the robust M-estimators of linear regression~\cite{Hampel1986,Huber2011} and is robust against outliers compared with the least-squares estimation of linear regression coefficients. It is also robust against violations of standard assumptions on the usual mean regression, such as heavy-tailed noise and skewed conditional and noise distributions.

In information geometry, a model manifold is often constructed  using a parametric distribution. 
Estimates are regarded as the projection of an empirical distribution onto the model manifold. In the case of linear regression, we construct a model manifold under the assumption that an error variable has a normal distribution. Because of the lack of a parametric distribution, constructing a model manifold that corresponds to the MLR model is difficult with conventional approaches. Some studies have considered nonparametric models for information geometry. Pistone and Sempi~\cite{Pistone1995} showed a well-defined Banach manifold for probability measures. Grasselli~\cite{Grasselli2010} addressed the Fisher information and $\alpha$-connections for the Banach manifold. Zhang~\cite{Zhang2013} discussed the relationship between divergence functions, the Fisher information, $\alpha$-connections, and fundamental issues in information geometry. 
In contrast to these nonparametric approaches to information geometry, in~\cite{DBLP:conf/iconip/SandoAMH18,Sando2019}, the geometric operation, which leads to the mode or the operation that makes the MLR estimator robust, was elucidated and an information geometric perspective on MLR was obtained. 

Let $x \in \mathbb{R}^p$ and $y \in \mathbb{R}$ be a set of predictor variables and a response variable, respectively. The original least squares for linear regression estimates a conditional mean of $y$ given $x$, while MLR estimates a conditional mode of $y$ given $x$. We briefly explain the EM algorithm of MLR introduced in~\cite{Yao2012}.

\subsubsection{Formulation}\label{sec:MLR_Formulation}
Suppose that $\left\{x_i, y_i \right\}_{i=1}^{N}$ are i.i.d. observations, where the $i$-th predictor variable is denoted by $x_i\in \mathbb{R}^p$ and the corresponding response is denoted by $y_i \in \mathbb{R}$. With MLR, a conditional mode of $y$ given $x$ by a linear function of $x$ is modeled as
\begin{align*}
  \text{Mode} \left[ y;x \right]
    &= x^{\top}\beta,
\end{align*}
where $\text{Mode} \left[ y;x \right] = \argmax_{y} f(y|x)$ for the conditional density function $f(y|x)$. 
Namely, $y$ and $x$ are related as
\begin{align}
  \label{eq:MLR_Formulation_model}
  y
    =  x^{\top}\beta + \epsilon, \quad 
\text{where} \quad \text{Mode} \left[ \epsilon;x \right] = 0.
\end{align}
To estimate $\beta$, Lee~\cite{Lee1989} introduced a loss function with the form
\begin{equation}
l(\beta ; y, x) =  - \phi_{h}
\left(
y - x^{\top}\beta
\right),
\end{equation}
where $\phi_h(x) = \frac{1}{h}\phi \left(\frac{x}{h} \right)$, $\phi(\cdot)$ is a kernel function, and $h$ is a bandwidth parameter. Minimizing the empirical loss leads to the estimate $\hat{\beta}$ of the linear coefficient:
\begin{align}
\hat{\beta}=  \argmax_{\beta} \frac{1}{N} \sum_{i=1}^{N} \phi_h(y_i - x_i^{\top}\beta).
  \label{eq:MLR_Formulation_problem}
\end{align}
In this paper, $\phi(\cdot)$ denotes a standard normal density function. The consistency and asymptotic normality of the estimate $\hat{\beta}$ obtained by Eq.~\eqref{eq:MLR_Formulation_problem} have been established under certain regularity conditions on the samples, kernel function, parameter space, and vanishing rate of the bandwidth parameter~\cite{Gordon2012}.

\subsubsection{EM algorithm for MLR}\label{sec:MLR_EMalgorithm}
Here, we introduce the EM algorithm for the MLR parameter estimation proposed in~\cite{Yao2012}. The algorithm consists of two steps starting from an initial estimate $\beta^{(1)}$.
\begin{description}[style=nextline]
  \item[E-Step:]
    Consider the surrogate function
        \begin{equation}
\gamma (\beta;\beta^{(k)}) =   \sum_{i=1}^{N} \pi_i^{(k)} \log \left[ \frac{ \frac{1}{N}\phi_h \left( y_i-x_i^{\top}\beta \right) }{\pi_i^{(k)}} \right],
      \label{eq:MLR_E-Step_surrogate}
      \end{equation}
where
    \begin{align}
      \pi_i^{(k)}
        = \frac{\phi_h(y_i - x_i^{\top}\beta^{(k)}) }{\sum_{j=1}^{N} \phi_h(y_j - x_j^{\top}\beta^{(k)}) },\quad
      i = 1 \dots N.
      \label{eq:MLR_E-Step}
\end{align}
 This function satisfies
\begin{equation}
\gamma (\beta^{(k)};\beta^{(k)}) =\log \left[ \frac{1}{N}\sum_{i=1}^{N}\phi_h\left( y_i - x_i^{\top}\beta^{(k)} \right) \right]
      \label{eq:MLR_E-Step_cond2}
\end{equation}
and
\begin{align}
      \log \left[ \frac{1}{N}\sum_{i=1}^{N} \phi_h \left( y_i-x_i^{\top}\beta \right) \right]
        &=  \log \left[ \sum_{i=1}^{N} \pi_i^{(k)} \frac{\frac{1}{N}\phi_h \left( y_i-x_i^{\top}\beta \right) }{\pi_i^{(k)}} \right]
    ,\quad
      \text{by Jensen's inequality}
      \notag
    \\
      &\geq \sum_{i=1}^{N} \pi_i^{(k)} \log \left[ \frac{ \frac{1}{N}\phi_h \left( y_i-x_i^{\top}\beta \right) }{\pi_i^{(k)}} \right]
      = \gamma (\beta;\beta^{(k)}).
     \label{eq:MLR_E-Step_cond1}
    \end{align}

  \item[M-Step:]
  In this step, the parameter $\beta$ is updated to increase the value of $\frac{1}{N}\sum_{i=1}^{N}\phi_h \left(y_i-x_i^{\top}\beta\right)$. The updated parameter $\beta^{(k+1)}$ is given as
    \begin{align}
      \beta^{(k+1)}
        &=  \argmax_{\beta} \gamma (\beta;\beta^{(k)})
      \label{eq:MLR_M-Step_optimize_surrogate}
    .
    \end{align}
 The following inequality holds:
    \begin{align*}
      \log \left[ \frac{1}{N}\sum_{i=1}^{N}\phi_h \left( y_i-x_i^{\top}\beta^{(k+1)} \right) \right]
        &\geq  \gamma (\beta^{(k+1)};\beta^{(k)})
    \\
        &\geq  \gamma (\beta^{(k)};\beta^{(k)})
        = \log \left[ \frac{1}{N}\sum_{i=1}^{N}\phi_h \left( y_i-x_i^{\top}\beta^{(k)} \right) \right]
    .
    \end{align*}
    Equation~\eqref{eq:MLR_M-Step_optimize_surrogate} is equivalent to
    \begin{align}
      \beta^{(k+1)}
        &=  \argmax_{\beta} \sum_{i=1}^{N} \pi_i^{(k)} \log \phi_h(y_i - x_i^{\top}\beta).
      \label{eq:MLR_M-Step}
      \end{align}
When $\phi(\cdot)$ is a standard normal density function,
\begin{align*}
\beta^{(k+1)}
        &=  \left( X^{\top} W_{k} X \right)^{-1} X^{\top} W_{k} y,
\quad
        &  W_{k} = \text{diag} \begin{pmatrix} \pi_1^{(k)} & \cdots & \pi_N^{(k)} \end{pmatrix}.
    \end{align*}
\end{description}
The property of the estimate $\hat{\beta}$ was discussed in~\cite{Yao2012}.

\subsubsection{Information geometry of MLR}\label{sec:informationgeometryofMLR}
Sando et al.~\cite{Sando2019} analyzed MLR from the viewpoint of information geometry. They elucidated the source of the difficulty by constructing a model manifold and data manifold for the MLR model and proposed a framework for geometrically formulating the MLR model.

To elucidate the cause of the difficulty of constructing manifolds for the MLR model,  consider the parameter estimation of a Gaussian mixture model as a specific example of statistical inferences in information geometry. Suppose that observations $x_i\in\mathbb{R}^{p}, i=1, \cdots, N$ are i.i.d. subject to a Gaussian mixture distribution expressed as
\begin{align*}
  f(x; \mu,\Sigma)
    &=  \sum_{i=1}^{K} \pi_i g(x;\mu_i, \Sigma_i)
,\\
    &\text{where} \quad \left\{
      \begin{aligned}
        &\pi_i \geq 0, \; \sum_{i=1}^{K} \pi_i = 1
      ,\\
        &g(x;\mu_i, \Sigma_i)
          = \frac{1}{\sqrt{2\pi}^p\sqrt{\mathrm{det}(\Sigma_i) }}
            \exp \left\{ -\frac{1}{2} (x-\mu_i)^{\top} \Sigma_i^{-1} (x-\mu_i) \right\}
      .
      \end{aligned}
    \right.
\end{align*}
Then, the model manifold consists of Gaussian mixture density functions whose parameters are the means and covariance matrices. The data manifold is constructed based on the empirical density function $\frac{1}{N}\sum_{i=1}^{N}\delta(x-x_i)$.

In the parameter estimation of the Gaussian mixture model, the model manifold is constructed on the basis of parametric distribution. In contrast, even though they have the similarity that densities are approximated by a mixture of kernel functions, there is no assumption of parametric distributions in MLR. This makes it nontrivial to construct a model manifold and data manifold.

To construct the model manifold for the MLR model,  consider (i) the assumption that $\text{Mode}\left[ \epsilon;x \right] = 0$ and (ii) the form of the objective function of $\beta$ for the MLR model: $\frac{1}{N} \sum_{i=1}^{N} \phi_h \left(y_i - x_i^{\top}\beta \right)$. From this assumption and fact, the optimization problem expressed in Eq.~\eqref{eq:MLR_Formulation_problem} is regarded as a maximization problem of KDE at $\epsilon = 0$ for the probability density function of $\epsilon$.
On the basis of the given observations, we propose constructing the following model for MLR:
\begin{align}
  f(\epsilon;\beta)
    &=  \frac{1}{N} \sum_{i=1}^{N} \phi_h \left( \epsilon - \epsilon_i(\beta) \right)
  \label{eq:MLRandIG_model}
,
\end{align}
where $\epsilon_i(\beta) = y_i - x_i^{\top}\beta,\ i=1\dots N$ and the variable $\epsilon$ denotes an error variable. In~\cite{AMARI19951379}, the latent variable $Z\in \left\{1\dots N \right\}$, which specifies a mixture component from which an observation is obtained, was introduced. The joint density function of $\epsilon$ and $Z$ is
\begin{align}
\label{eq:jointD}
  g(\epsilon,z;\beta)
    &=  \prod_{i=1}^{N} \left[ \frac{1}{N} \phi_h \left( \epsilon - \epsilon_i(\beta) \right) \right]^{\delta_i(z)},
\end{align}
where $\delta_i(z) = 1$ if $z=i$ and $\delta_i(z)=0$ if $z\neq i$. 
The model manifold $\mathcal{M}$ is denoted by
\begin{align}
  \mathcal{M}
    &=  \left\{ g(\epsilon,z;\beta) \mid \beta \in \mathbb{R}^{p} \right\},
  \label{eq:MLRandIG_modelmanifold}
\end{align}
which is a curved exponential family.

We next consider constructing a data manifold for the MLR model. The empirical density function is often constructed on the basis of observations. 
The empirical density function is constructed as follows:
\begin{align}
  p(\epsilon)
    &=  \delta(\epsilon - 0)
    =   \delta(\epsilon)
  \label{eq:MLRandIG_empiricaldensityfunction}
.
\end{align}
By introducing the latent variable $Z\in \left\{1\dots N \right\}$ and the parameters $\left\{q_i \right\}_{i=1}^{N} (q_i \geq 0, \sum_{i=1}^{N} q_i = 1)$ to Eq.~\eqref{eq:MLRandIG_empiricaldensityfunction}, 
$p(\epsilon)$ is extended to the empirical joint density function of $\epsilon$ and $Z$:
\begin{align}
  h(\epsilon, z\ ; q_1 \dots q_N)
    &=  \sum_{i=1}^{N} q_i \delta(\epsilon) \delta_i(z)
  \label{eq:MLRandIG_empiricaljointdensityfunction}
,\quad \text{where} \quad q_i \geq 0, \;\sum_{i=1}^{N} q_i = 1
\end{align}
The data manifold $\mathcal{D}$ is defined as
\begin{align*}
  \mathcal{D}
    = \left\{
        h(\epsilon,z\ ;q_1\dots q_N)
      \mid
        q_i \geq 0,
      \quad
        \sum_{i=1}^{N} q_i = 1
      \right\}
.
\end{align*}
$\mathcal{D}$ is shown to be a mixture family.

Consider the $e$-projection of a model with the parameters $\beta^{(k)}$ onto the data manifold:
\begin{align}
  &\min_{h \in \mathcal{D}} D(h,g(\cdot,\cdot;\beta^{(k)})),
\end{align}
which is equivalent to the problem
\begin{align}
      \min_{q_1 \dots q_N}
        \quad D \left(h(\cdot,\cdot;q_1 \dots q_N),g(\cdot,\cdot;\beta^{(k)}) \right)
    \quad
      \text{s.t.}
            \quad  q_i \geq 0, \; \;\sum_{i=1}^{N} q_i = 1.
\label{eq:MLRandIG_optimizeQ}        
\end{align}
An optimal solution for Eq.~\eqref{eq:MLRandIG_optimizeQ} is
\begin{align}
  q_i^{(k)}
    = \frac{\phi_h \left( y_i - x_i^{\top}\beta^{(k)} \right) }{\sum_{j=1}^{N} \phi_h \left( y_j - x_j^{\top}\beta^{(k)} \right) },
\quad
  i  = 1 \dots N,
  \label{eq:MLRandIG_optimalQ}
\end{align}
which is equivalent to the E-step in Eq.~\eqref{eq:MLR_E-Step}.
Then, consider the $m$-projection of the empirical joint density function with the parameters $q_i=q_i^{(k)},\ i=1\dots N$ onto the model manifold:
\begin{align*}
  \min_{g \in \mathcal{M}} D(h(\cdot,\cdot\ ;q_1=q_1^{(k)}\dots q_N=q_N^{(k)}),g)
,
\end{align*}
which is solved by
\begin{align}
  \max_{\beta} \sum_{i=1}^{N} q_i^{(k)} \log \phi_h \left( y_i - x_i^{\top}\beta \right),
  \label{eq:MLRandIG_equiv_optimizebeta}
\end{align}
and is equivalent to the M-step~\eqref{eq:MLR_M-Step}.

\begin{figure}
\centering
  \includegraphics[width=.8\linewidth]{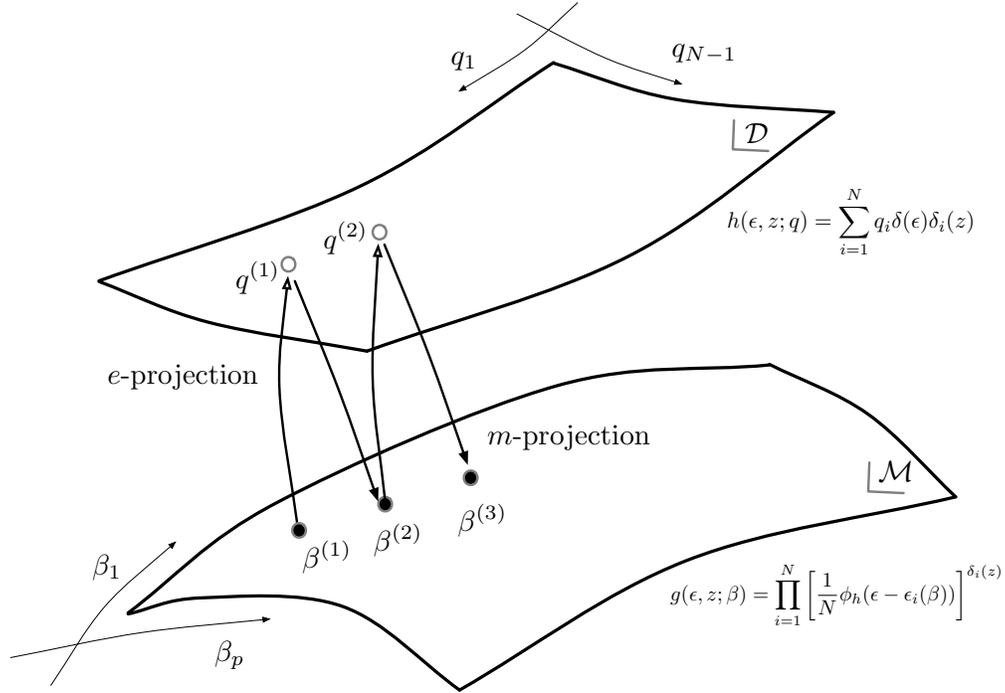}
  \caption{Conceptual diagram of the $em$ algorithm corresponding to the MLR model}
  \label{fig:MLRandIG_diagramoftheemalgorithmofmlr}
\end{figure}
Figure~\ref{fig:MLRandIG_diagramoftheemalgorithmofmlr} shows the update process of the $em$ algorithm corresponding to the MLR model parameter estimation.

\subsection{Boltzmann machine learning}
The Boltzmann machine (BM) is a fully connected neural network model with $n$ neurons and is equivalent to a second-order log linear model with a binary random variable of length $n$, and is trained to approximate the distribution of the input data. A Boltzmann machine is usually composed of $v$ visible units $x = (x_1,\dots,x_v) \in \{0,1\}^{v}$ and $h$ hidden units $y = (y_1,\dots,y_h) \in \{0,1\}^{h}$. 
The family of Boltzmann machines with $n = v+h$ neurons is written as 
\begin{align}
\cB = \left\{
B \in \cP^{n} :
B(z) = b \exp 
\left\{
\sum_{1 \leq i < j \leq n} w_{ij} z_i z_j
\right\}
\right\},
\end{align}
where $w_{ij}, 1 \leq i,j \leq n$ are the network connectivities, $z=(x,y)$ and $\cP^n$ is the collection of probability distributions on the set of machine states $\{0,1\}^{n}$. The constant $b$ is a normalizing factor so that for all $w_{ij}$, the function $B(z)$ is a valid distribution. The behavior of the $v$ visible units is described by the marginal distribution $B_v$ determined from $B$ as 
\begin{align}
    B_v(x) = \sum_{y \in \{0,1\}^h} B((x,y)).
\end{align}
The objective of training the BM is to find a machine for which $B_v$ is close to the empirical distribution $\hat{P}$ of the observation of visible units.  Namely, by defining the family $\cD$ of the desirable distributions on the set of machine state $\{0,1\}^n$ as 
\begin{align}
    \cD = 
    \left\{
    P \in \cP^{n} : 
    \sum_{y \in \{0,1\}^{h}} P((x,y)) = \hat{P}(x),\; \forall x \in \{0,1\}^{v}
    \right\}
\end{align}
whose marginal distribution on the visible units agrees with $\hat{P}$, the problem of BM learning is formulated as
\begin{align}
    \inf_{B \in \cB} \inf_{P \in \cD} D(P , B).
\end{align}
Since $\cD$ is a subspace of distributions consistent with the given observed (visible) variable, it plays a similar role to the data distribution in terms of the $em$ algorithm. Detailed convergence analysis of this bilevel optimization problem is given in~\cite{143375}. 

It is shown that the manifold of BM without hidden units is an $e$-flat manifold, so that an invariant metric is introduced into it. 

It is shown that the best approximation is given by the $m$-geodesic projection, and is unique in the case of no hidden units. Furthermore, a generalized Pythagorean theorem makes it possible to decompose the approximation error in an invariant manner. The BM manifold with hidden units is not $e$-flat but has some interesting properties~\cite{125867}. The possibility of further speed up by simultaneously solving both the $e$- and $m$-steps using information geometrically derived gradient flows is suggested in~\cite{FUJIWARA1995317}. 

Also, an information geometric structure of Helmholtz machine learning, known as the Wake--Sleep algorithm, was elucidated in~\cite{NIPS1998_0771fc6f}, where, in contrast to the EM and $em$ algorithms, both the Wake-phase and Sleep-phase correspond to the $m$-projection.

\section{Data analysis for distributional data}

Here, we regard a distribution as a datum.
In other words, a point $\theta$ in the parameter space $\mathcal{S}$ is given as a datum.
Such generalization of data analysis in the Riemannian manifold has been attracting much attention~\cite{DBLP:journals/tmi/FletcherLPJ04}.
One example of such a situation is found in the field of sensor fusion, where numerous sensors are distributed and plenty of data are obtained by each sensor, 
but a high communication capacity is required to collect all data from all sensors.
One way to reduce the communication cost is to collect only a distribution parameter
calculated in each sensor.
Another example is transfer learning in machine learning, where there are many different tasks, each of which
includes a set of data. By using such source tasks, it may be possible to improve the accuracy for a new
target task that only has a small number of samples. 

In transfer learning, it is necessary to model common features among different task
at the same time as the difference between features.
Although there are various ways of formulation of transfer learning\cite{zhuang2020comprehensive}, 
we assume that all tasks lie on a lower dimensional latent subspace representing a common structure
among tasks and the difference between tasks is
expressed as different locations on the subspace\cite{pan2008transfer}.

In this section, we explain the EM-like iterative algorithm used in such a situation. 

\subsection{Statistical inference for distributional data}

Let us consider a transfer learning scenario.
Suppose we have distributions $p_1,\ldots,p_N$ from source tasks and
$p_{\new}$ for a target task, where $p_{\new}$ is only based on a small number of samples.
One simple way of transfer learning is to find a projection from $p_{\new}$
onto a flat subspace spanned by $p_1,\ldots,p_N$.
If we take the $e$-flat subspace, the subspace 
is 
\begin{equation}
    \mathcal{M}_e = \left\{\theta \;\middle|\; \theta = \sum_i w_i \theta_i, \sum_i w_i=1, \theta\in \mathcal{S}\right\},
\end{equation}
where $\theta_i$ is an $e$-coordinate of $p_i$, and we can define the $m$-flat subspace similarly.
This transfer learning is merely a simple projection, and it can be solved by calculating
the projection point (for uniqueness, it is natural to take $m$-projection for an $e$-flat subspace and
$e$-projection for an $m$-flat subspace), which can be obtained in an explicit form or by a gradient descent method.
However, the problem becomes difficult when
$p_i$ is given by an empirical distribution $p_i(x) = (1/N_i)\sum_j \delta(x-x_{ij})$, where $\mathcal{X}_i=\{x_{ij}\}$ is a sample set for $p_i$, and $N_i=|\mathcal{X}_i|$. 
In such a nonparametric case, the distribution does not have an explicit form of the $e$-coordinate.
Takano et al.~\cite{DBLP:journals/neco/TakanoHAM16} solved this difficulty by avoiding the explicit
expression of the $e$-coordinate and introducing a new geometrical algorithm based on a generalized Pythagorean theorem, as described below.

Instead of an explicit form of the $e$-coordinate, we can use a characteristic of an $e$-mixture that is a member of $\mathcal{M}_e$, shown by Murata and Fujimoto~\cite{murata2009bregman}
\begin{equation}
\label{eq:emixture}
    p_w = \argmin_{q\in\mathcal{S}} \sum_i w_i D(q, p_i),
\end{equation}
where $p_w$ is an $e$-mixture defined as a distribution whose $e$-coordinate is $\sum_i w_i \theta_i$ in a parametric case.
The right-hand side can be used as an implicit expression of the $e$-mixture, which
only depends on the weight $w_i$ and divergence between $q$ and $p_i$.
To find the projection of $p_{\new}$ onto $\mathcal{M}_e$, we must determine $q$ and $w_i$ in \eqref{eq:emixture}.
Takano et al.~proposed an algorithm for optimizing $q$ and $w_i$ alternatively, as in the EM algorithm.
Instead of the $e$-coordinate, $q$ is expressed in a nonparametric form, 
$q(x) = \sum_j v_j \delta(x-y_j)$, where the set $\mathcal{Y} = \{y_j\}$ is typically 
$\mathcal{Y}\subset\bigcup_i \mathcal{X}_i$.

If $w_i$ is fixed, the right-hand side of \eqref{eq:emixture} is minimized with respect
to $v_j$. This can be performed if $D(q, p_i)$ is expressed as a function of $v_j$ and
$x_{ij}$. Such a nonparametric expression of divergence has been proposed in the research
field of point process, 
for example, the method proposed in~\cite{hino2013information,hino2015non}. By using gradient descent of the expression, $v_j$ is optimized for fixed $w_i$.

On the other hand, the projection of $p_{\new}$ onto $\mathcal{M}_e$ must satisfy
the orthogonality
\begin{equation}
    D(p_{\new}, p_i) = D(p_{\new}, q) + D(q, p_i)
\end{equation}
for all $i$.
However, the current values of $w_i$ and $v_j$ do not necessarily satisfy this formula. Takano et al.\ proposed a simple update rule for $w_i$ as follows:
if the left-hand side is larger than the right hand side, this means the point $q$
is too close to $p_i$, hence $w_i$ should be decreased from the current value; on the other hand, if the left-hand side is smaller, $w_i$ should be increased.
The convergence of this algorithm has been investigated by Akaho et al.\cite{DBLP:conf/iconip/AkahoHM19}, who showed that the algorithm is guaranteed to converge under a mild condition.

\subsection{Dimension reduction for distributional data}
\begin{figure}[ht]
\centering
  \includegraphics[width=.6\linewidth]{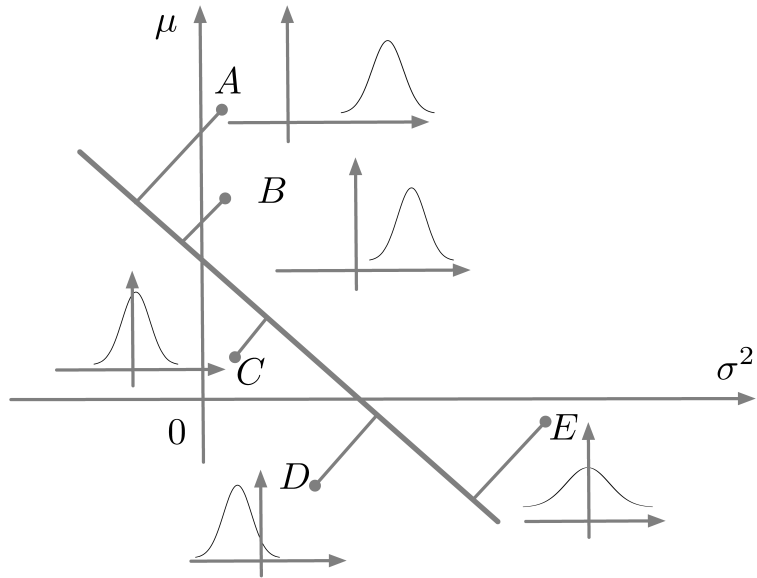}
  \caption{Schematic figure of PCA for the parameters of a distribution.}
  \label{fig:ePCA}
\end{figure}

The method above becomes difficult if the number of source tasks increases, because
the dimension of subspace becomes large, which may cause a curse of dimensionality.
To reduce the dimension, the most well-known method is the principal component analysis (PCA). PCA finds the subspace that minimizes the mean square distance from sample points, which implicitly assumes that the sample points lie on the Euclidean space.
Although PCA can be applied even to the sample points that are distribution parameters (Fig.~\ref{fig:ePCA}), there are some serious issues.
One is that the projection from the point to the subspace is not necessarily included in the
domain, for instance, it can be negative variance for Gaussian distribution as in the example in Fig.~\ref{fig:ePCA}.
Another is that the Euclidean distance between distribution parameters would not be appropriate from a statistical point of view.

From such considerations, Akaho~\cite{akaho2004pca} and  Collins et al.\cite{collins2001generalization} proposed an extension of PCA to the case of probability distributions, in particular, exponential family distributions.
As described in the previous sections, exponential family distributions have two dual coordinates, $e$-coordinate
$\theta$ and $m$-coordinate $\eta$, and for each coordinate, the flat subspace is given by a linear equation of the coordinate.
By this duality, there are two kinds of extension of PCA: $e$-PCA, which finds the flat subspace for the $e$-coordinate, and
$m$-PCA for the $m$-coordinate.

We describe the $e$-PCA here, while the $m$-PCA is defined in a similar way merely by exchanging $e$- with $m$- below.
The goal of the $e$-PCA is to determine an affine subspace defined using $u=u_1,\ldots,u_K$ for some fixed $K$,
\begin{equation}
    \mathcal{M}_e(u) = \left\{\hat{\theta}\;\middle|\; \hat{\theta}=\sum_{k=1}^{K} w_k u_k \in\mathcal{S}, \sum_{k=1}^{K} w_k=1\right\}
\end{equation}
to fit given samples $\theta_1,\ldots,\theta_N$.
To guarantee a unique projection from a sample point to the subspace, it is natural to take
dual projection, i.e., $m$-projection for $e$-PCA and $e$-projection for $m$-PCA. The projection can be formulated in terms of KL divergence, therefore the objective function for $e$-PCA is
\begin{equation}
    L(w, u) = \sum_{i=1}^N D(\theta_i, \hat{\theta}_i),\quad \hat{\theta}_i = \sum_{k=1}^{K} w_{ik} u_k,
\end{equation}
where $\hat{\theta}_i$ is the projection point of $\theta_i$.
We need to optimize this function with respect to $w_{ik}$ and $u_k$. If we fix $u_k$, the weights $w_{ik}$ are coefficients for the projected points of $\theta_i$, which can be
uniquely determined from the duality, and it is typically minimized by a gradient descent method.
The first derivative of $L(w, u)$ is given in a simple form,
\begin{equation}
    \frac{\partial L(w, u)}{\partial w_{ik}} = u_k (\hat{\eta}_i - \eta_i),\quad
    \frac{\partial L(w, u)}{\partial u_k} = \sum_{i=1}^{N} w_{ik} (\hat{\eta}_i - \eta_i),
\end{equation}
where $\eta_i$ and $\hat{\eta}_i$ are the $m$-coordinates of $\theta_i$ and $\hat{\theta}_i$ respectively.
From the above equations, we can optimize $w_{ik}$ and $u_k$ alternatively.
Geometrically, it is easy to see that optimizing $w_{ik}$ for a fixed $u_k$ is the $m$-projection from $\theta_i$ onto a fixed subspace $\mathcal{M}_e(u)$.
It also holds that optimizing $u$ is the $m$-projection of $(\theta_1,\ldots,\theta_N)$ 
onto a fixed subspace $\mathcal{N}_e(w)\subset \mathcal{S}^N$, where
\begin{equation}
\mathcal{N}_e(w)=\left\{(\hat{\theta}_1,\ldots,\hat{\theta}_N)\;\middle|\; \hat{\theta}_i=\sum_{k=1}^{K} w_{ki} u_k, \hat{\theta}_i\in\mathcal{S}\right\}.
\end{equation}
Therefore, both alternating optimizations of $w$ and $u$ have a unique optimum, but it does not mean that
the algorithm finds the global optimum as in the case of the EM algorithm.

\subsection{Related topics on the dimension reduction for distributional data}

The simplest application of $e$-PCA or $m$-PCA would be the case of $K=1$, which is a general center of samples.
The $e$-center, that is, the $e$-PCA of this case, is explicitly given by 
\begin{equation}
    u_1=\theta\left(\frac1N\sum_{i=1}^{N}\eta_i\right), 
\end{equation}
where $\theta(\eta)$ is the coordinate transformation of $\eta$ to the $e$-coordinate, and $\eta_i$ is the $m$-coordinate of $\theta_i$.
The $m$-center has a similar form by exchanging $\theta$ and $\eta$.
By using the $e$-center or $m$-center, we can generalize the k-means clustering to distributional data in a natural way.
Watanabe et al.\cite{watanabe2009variational} proposed the method for solving clustering and dimension reduction simultaneously, where
Bayesian formulation is also introduced.

The nonnegative matrix factorization (NMF)\cite{lee2000algorithms,cichocki2009nonnegative}
is also a widely used dimension reduction method. 
Given a high-dimensional data matrix $X$ with all nonnegative elements, we consider an approximation of $X$ by 
$W H$, where $W$ and $H$ are lower rank matrices with all nonnegative elements.
Supposing $X=WH$ holds, let us consider the matrix $\hat{X}$, which is a normalization of $X$ so that the sum of elements of each
row is equal to 1.
We can show that there exist $\hat{W}$ and $\hat{H}$ such that $\hat{X}=\hat{W}\hat{H}$,
where $\hat{W}$ and $\hat{H}$ are also normalized.
Since nonnegative values summing to 1 can be regarded as a probability vector, the rows of $\hat{X}$, $\hat{W}$, and $\hat{H}$
can also be regarded as a probability vector.
On the basis of this fact, Akaho et al.~\cite{akaho2018geometrical} provided a geometrical view of NMF.
The probability vector is an $m$-coordinate of multivariate distribution, and the NMF can be considered as
a dimension reduction of distributional data.
This is similar to the $m$-PCA of the probability vectors, the difference being that the decomposed matrix must be nonnegative,
which is a stronger assumption than that in the $m$-PCA. Without loss of generality, all values $u_k$ can be positive in the $m$-PCA, and NMF is the restriction of the $m$-PCA so that $w_{ik}\ge0$ is satisfied.
In the NMF, the maximum likelihood estimation is often applied, which is equivalent to the $m$-projection.
Although it is more natural to take the $e$-projection for the $m$-flat subspace, $m$-projection is also unique in this special case.

Another example is dimension reduction for the Gaussian process. 
The Gaussian process is a stochastic process on some set $\mathcal{X}$, 
which is defined by a mean function $m(x)$ and a covariance function $v(x,x')$, where
$x,x'\in\mathcal{X}$. For any set of values $x_1,\ldots,x_M\in\mathcal{X}$, function values $f_1,\ldots,f_M$ are generated from a multivariate Gaussian
distribution with mean $(m(x_1),\ldots,m(x_M))$ and covariance matrix $V=(v(x_i,x_j))_{i,j=1,\ldots,M}$.
Since the number of points $M$ is arbitrary, the Gaussian process is essentially an infinite dimensional distribution.
Supposing there is a set of $N$ Gaussian processes $G\!P_1,\ldots,G\!P_N$, we can consider the application of the $e$-PCA or $m$-PCA to those 
Gaussian processes. However, it is not trivial owing to its infinite-dimensional nature.
Ishibashi and Akaho~\cite{ishibashi2022principal} proved that if the Gaussian processes are different posteriors of the same prior distribution, the infinite-dimensional
$e$-PCA ($m$-PCA) can be reduced to the finite-dimensional $e$-PCA ($m$-PCA). They also provided some applications to transfer learning.

We have described the application of the $e$-PCA or $m$-PCA limited to the case that the distributions belong to the exponential family.
One possible extension is a mixture distribution, $p(x;q,\theta)=\sum_{k=1}^{K} q_k p_k(x; \theta_k)$, which is often used for clustering. 
Here, we consider the mixture of the exponential family,
\begin{equation}
    p(x) = \sum_{k=1}^{K} \pi_k f_k(x;\xi_k), \qquad \sum_{k=1}^K \pi_k = 1, \qquad \pi_k\ge0
\end{equation}
where $\pi_k$ is the weight parameter for a component distribution $f_k(x;\xi_k)$.
This distribution does not belong to the exponential family even when $f_k$ is an exponential family distribution.
However, if we introduce a latent variable $z$ such that input variable $x$ is generated from $f_z(x;\xi_k)$,
the joint distribution of $(x,z)$, $p(x,z) = \pi_z f_z(x;\xi_z)$ becomes an exponential family, which means that the mixture of exponential family distributions can be embedded into the space of the exponential family.
On the basis of this idea, Akaho~\cite{akaho2008dimension} extended the framework of $e$-PCA and $m$-PCA to the case of a mixture of exponential family distributions.
The key issue is the freedom of the order of components, i.e., the permutation of $z$ gives a different embedding
even though the resulting mixture distributions are identical.
Supposing two mixtures of distributions $p_1$ and $p_2$ written in latent variable forms,
\begin{equation}
    p_1(x, z)=a_z f_z(x;\xi_z), \qquad p_2(x, z)=b_z f_z(x;\zeta_z),  
\end{equation}
the divergence between $p_1$ and $p_2$ gives different values, depending on the order of $z$. 
Akaho resolved this problem by optimizing the order of $z$ so as to minimize the KL divergence between $p_1(x,z)$ and $p_2(x,z)$.
This problem can be solved by the linear programming method.

\section{Neural generative model}

An image is represented as a single point in a high-dimensional space,
but not every point in this space corresponds to a ``natural image''.
For example, if randomly generated high-dimensional data were displayed in the same format as an image, it would look like a ``sand storm'' and would not be meaningful as an image.
In other words, natural images occupy only a small region of the high-dimensional
space, and this region might be relatively small compared with the entire space. To handle this problem in a probabilistic way, let us assume that the natural image is generated in accordance with a probability distribution $P$ in the higher-dimensional space, with some regions assigned a high probability and other regions assigned a very low, even $0$, probability. 
If we can well represent this probability distribution, which might be condensed at a very small region in high-dimensional space,
we can generate a natural image stochastically.
The idea of generative adversarial nets~(GAN; \cite{NIPS2014_5ca3e9b1})
is introduced as a method for modeling natural
images using a deep neural network.

Let $\mathcal{X}$ denote the high-dimensional space containing images. Let $P$ be the probability distribution on $\mathcal{X}$ that generates the natural image, and let $\mathcal{D}=\{x_{1},\dotsc,x_{n}\}$ be a data set of natural images sampled from $P$.
We also prepare a low-dimensional probability distribution $R$ with known characteristics.
For example, a uniform distribution on $[0,1]^{k}$ or a normal distribution on $\mathbb{R}^{k}$ is often used.
Let $\mathcal{Z}$ be  this sample space and $R$ be the probability distribution on $\mathcal{Z}$.
We prepare a generator $G_{\theta}$, which is a machine that takes a single point on $\mathcal{Z}$ and generates a pseudo image on $\mathcal{X}$,
\begin{equation}
  z\in\mathcal{Z}\mapsto
  y=G_{\theta}(z)\in\mathcal{X},
\end{equation}
and the output can be changed depending on parameter $\theta$.
The generator $G_{\theta}$ is required to perform very complex transformations, so a deep neural network is commonly used. The pseudo image $Y$ is generated as
\begin{equation}
  Y=G_{\theta}(Z),\quad Z\sim R,
\end{equation}
therefore, the probability distribution of $Y$ is the probability distribution on $\mathcal{X}$ determined by the generator $G_{\theta}$
and the reference distribution $R$ on $\mathcal{Z}$,
which is called a pushforward measure.
A formal definition is as follows.
Let $\mathcal{X}$ be a sample space
and
$\mathcal{F}_{\mathcal{X}}$ be a $\sigma$-algebra on $\mathcal{X}$. 
Given a measurable function $G_{\theta}$,
the pushforward measure of $R$ is defined by
\begin{equation}
  P_{\theta}(B)=R(G_{\theta}^{-1}(B\cap G_{\theta}(\mathcal{Z}))),\;
  \forall B\in\mathcal{F}_{\mathcal{X}}.
\end{equation}
Note that the generator
$G_{\theta}$ only transforms the low-dimensional space $\mathcal{Z}$,
so that the support of $P_{\theta}$ is essentially a subspace of the same dimension with $\mathcal{Z}$.
To approximate the probability distribution of a natural image, the pseudo images generated by transforming the data set $\mathcal{C}=\{z_{1},\dotsc,z_{n}\}$ on $\mathcal{Z}$ from the reference distribution $R$
\begin{equation}
  \mathcal{\tilde{D}}
  =G_{\theta}(\mathcal{C})
  =\{y_{1},\dotsc,y_{n}\}
\end{equation}
imitate the natural images as if it comes from the same distribution of $\mathcal{D}$. 
This is an interesting and smart variation of the two-sample problem in classical statistics.
The difference between the two samples can be evaluated using an appropriate statistical distance,
for example the Jensen--Shannon divergence in the original work \cite{NIPS2014_5ca3e9b1},
$f$-divergences with variational lower bound optimization \cite{5605355} in \cite{NIPS2016_cedebb6e,NIPS2017_2f2b2656},
the maximum mean discrepancy associated with a certain reproducing kernel Hilbert space \cite{JMLR:v13:gretton12a} in \cite{pmlr-v37-li15,10.5555/3020847.3020875}, 
and the Wasserstein order 1 distance approximated by neural networks in \cite{pmlr-v70-arjovsky17a}.
In the following discussion, we focus on the original GAN procedure proposed in \cite{NIPS2014_5ca3e9b1}.

To optimize the generator $G_{\theta}$, Goodfellow et al.~\cite{NIPS2014_5ca3e9b1} proposed an adversarial procedure. They introduced a discriminative model $D_{\phi}$ as well as a generative model $G_{\theta}$ and alternately trained both models as a minimax game.
The learning procedure for the discriminative model is designed to estimate the probability that a sample comes from the distribution of natural images rather than pseudo images, which is given by
\begin{equation}
  \text{maximize }
  L(\phi)
  =
  \mathbb{E}_{X\sim P}[\log D_{\phi}(X)]
    +
    \mathbb{E}_{X\sim P_{\theta}}[\log(1-D_{\phi}(X))],
\end{equation}
where $\mathbb{E}_{X\sim P}$ stands for
the average with respect to the distribution $P$ and it is replaced by 
the empirical average with a data set $\mathcal{D}=\{x_{i},i=1,\dotsc,n\}$
sampled from a distribution $P$ in the learning process
\begin{equation}
  \mathbb{E}_{X\sim P}[f(X)]
  \to
  \frac{1}{n}\sum_{i=1}^{n}f(x_{i})
\end{equation}
and
$\mathbb{E}_{X\sim P_{\theta}}$ is also replaced 
with a data set $\mathcal{C}=\{z_{i},i=1,\dotsc,n\}$
from the reference distribution $R$ as
\begin{equation}
  \mathbb{E}_{X\sim P_{\theta}}[f(X)]
  \to 
  \frac{1}{n}\sum_{i=1}^{n}f(G_{\theta}(z_{i})).
\end{equation}
Also, the procedure for the generative model is designed
to maximize the probability
that the discriminative model makes a mistake,
which is given by
\begin{align}
  \text{minimize }
  L(\theta)
  &=
  \mathbb{E}_{X\sim P_{\theta}}[\log(1-D_{\phi}(X))]\\
  &=
  \mathbb{E}_{Z\sim R}[\log(1-D_{\phi}(G_{\theta}(Z)))].
\end{align}

To see the geometrical picture of this sophisticated procedure, we introduce two model manifolds.
One is a set of distributions of pseudo images generated by $G_{\theta}$,
\begin{equation}
  \mathcal{M}_{G}
  =
  \left\{P_{\theta} :
    \text{pushforward measure of $R$ with }
    G_{\theta},\;\forall\theta\in\Theta
  \right\},
\end{equation}
which corresponds to the generative model, and the other is a set of distributions for approximating midpoints of
the grand truth distribution $P$ and the generator's distribution $P_{\theta}$,
\begin{equation}
  \mathcal{M}_{D}
  =
  \left\{Q_{\phi} :
    \text{approximator of the $m$-midpoint of $P$ and $P_{\theta}$} \right\}
\end{equation}
which corresponds to the discriminative model.
The loss of $D$ for discriminating datasets from $P$ and $P_{\theta}$ is defined as
\begin{equation}
  L(D)=\mathbb{E}_{P}[\log(D(X))]+\mathbb{E}_{P_{\theta}}[\log(1-D(X))],
\end{equation}
therefore, the optimal discriminator $D_{*}$ is given by
\begin{equation}
  D_{*}(x)=\frac{p(x)}{p(x)+p_{\theta}(x)},
\end{equation}
where $p$ and $p_{\theta}$ are $m$-representations of $P$ and $P_{\theta}$, i.e., probability density functions of $P$ and $P_{\theta}$, respectively.
Hence the crucial part for discriminator learning can be regarded as the estimation of the midpoint of $P$ and $P_{\theta}$. 
Using these models,
the learning procedure is rewritten as follows.

\begin{figure}[ht]
\centering
  \includegraphics[width=.6\linewidth]{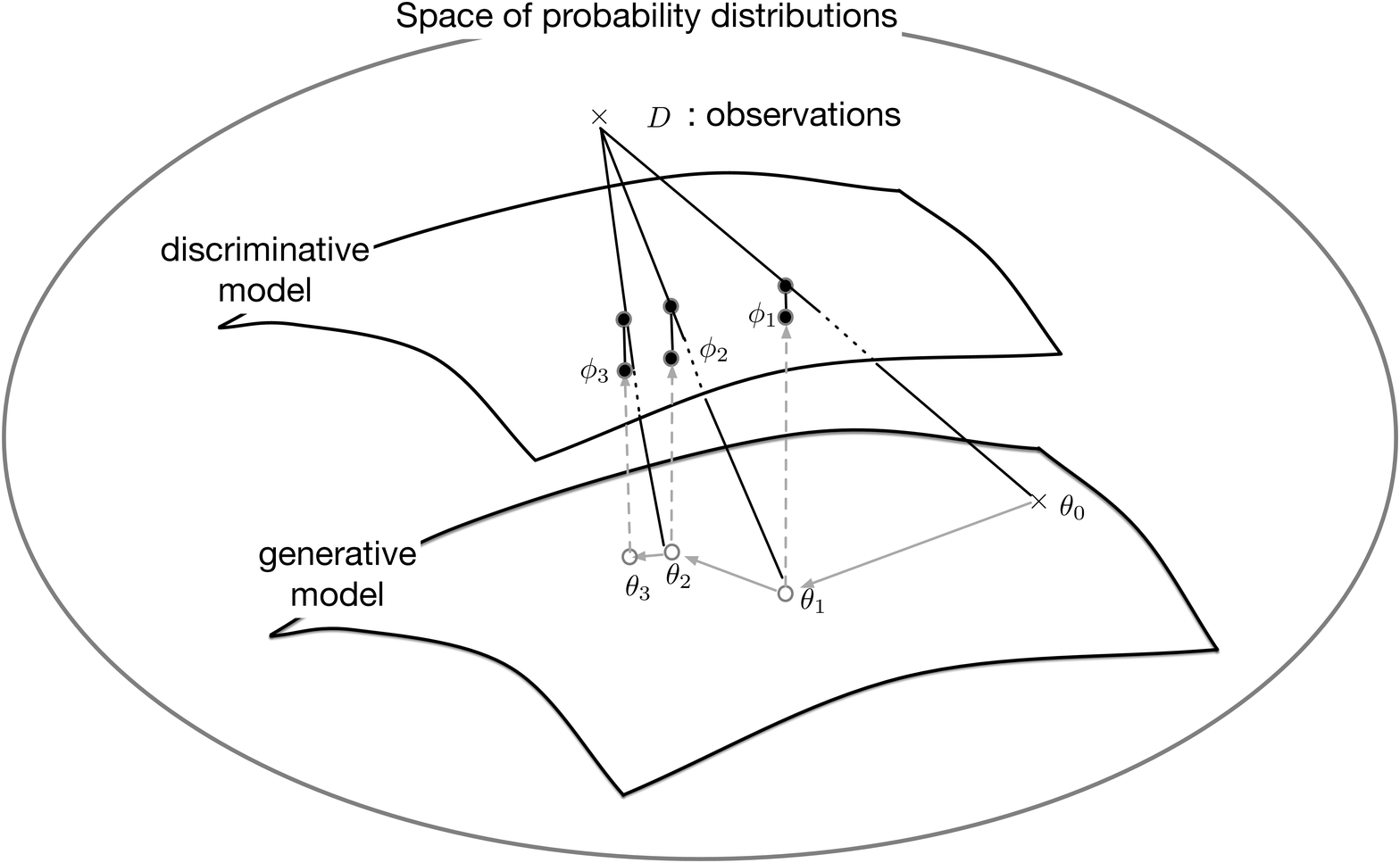}
  \caption{Schematic figure of generative adversarial model.}
  \label{fig:GAN}
\end{figure}

Let $q_{\phi}(x)$ be the estimate of
the midpoint $(p(x)+p_{\theta}(x))/2$ in
the discrimative model manifold $\mathcal{M}_{D}$.
Using the relation
\begin{equation}
  1-D(x)
  =
  1-\frac{p(x)}{p(x)+p_{\theta}(x)}
  =
  \frac{p_{\theta}(x)}{p(x)+p_{\theta}(x)},
\end{equation}
the corresponding reversal discriminator $1-D_{\phi}$ is
represented as
\begin{equation}
  1-D_{\phi}(x)
  =
  \frac{p_{\theta}(x)}{2q_{\phi}(x)}.
\end{equation}
Therefore, given a discriminator $D_{\phi}$,
the optimal parameter of the generative model is estimated as
\begin{align}
  \hat\theta
  &=\arg\min_{\theta}
    \mathbb{E}_{X\sim P_{\theta}}[\log(1-D_{\phi}(X))]\\
  &=\arg\min_{\theta}
    \mathbb{E}_{X\sim P_{\theta}}[\log p_{\theta}(X)-\log 2q_{\phi}(X)]\\
  &=\arg\min_{\theta}
    D(P_{\theta},Q_{\phi}),
\end{align}
which is the $e$-projection
from $Q_{\phi}$ on $\mathcal{M}_{G}$.
In the same way,
for a given generative model $G_{\theta}$,
the optimal parameter of the discriminative model is estimated as
\begin{align}
  \hat\phi
  &=\arg\max_{\phi}
    \mathbb{E}_{X\sim P}[\log D_{\phi}(X)]
    +
    \mathbb{E}_{X\sim P_{\theta}}[\log(1-D_{\phi}(X))]\\
  &=\arg\max_{\phi}
    \mathbb{E}_{X\sim P}[\log p(X)-\log 2q_{\phi}(X)]
    +
    \mathbb{E}_{X\sim P_{\theta}}[\log p_{\theta}(X)-\log 2q_{\phi}(X)]\\
  &=\arg\min_{\phi}
    \mathbb{E}_{X\sim P}[\log q_{\phi}(X)]
    +
    \mathbb{E}_{X\sim P_{\theta}}[\log q_{\phi}(X)]\\
  &=\arg\min_{\phi}
    \mathbb{E}_{X\sim (P+P_{\theta})/2}[\log q_{\phi}(X)]\\
  &=\arg\min_{\phi}
    D((P+P_{\theta})/2,Q_{\phi}),
\end{align}
which is the $m$-projection
from $(P+P_{\theta})/2$ on $\mathcal{M}_{D}$.
These two projections are iterated until convergence,
and their geometrical interpretation is schematically depicted
in Fig.~\ref{fig:GAN}.

It is worth noting that the adversarial procedure is
summarized in terms of the Jensen--Shannon divergence.
The Jensen--Shannon divergence 
is defined by using KL divergence with the $m$-midpoint as
\begin{equation}
  D_{JS}(P,P_{\theta})
  =
  D(P,(P+P_{\theta})/2)+
  D(P_{\theta},(P+P_{\theta})/2).
\end{equation}
We introduce an approximated version of
the Jensen--Shannon divergence with $Q_{\phi}$ as
\begin{equation}
  \tilde{D}_{JS}(P,P_{\theta};Q_{\phi})
  =
  D(P,Q_{\phi})
  +
  D(P_{\theta},Q_{\phi}).
\end{equation}
Then the original two-sample problem is formulated as
\begin{equation}
  \text{minimize }
  \tilde{D}_{JS}(P,P_{\theta};Q_{\phi})
  \text{ with respect to $\theta$ and $\phi$}.
\end{equation}

\section{Conclusion and extension}
We described the $em$ algorithm, the information geometric counterpart of the EM algorithm identified by Amari~\cite{AMARI19951379}. The $em$ algorithm is a meta-algorithm, which has a very wide range of applicability, and leaves room for customization to individual problems. Various extensions and applications of the $em$ algorithm were presented in this paper as  good examples of how viewing a problem from a geometric point of view clarifies the structure of the problem and facilitates parameter estimation using iterative algorithms. 

The EM algorithm was based on Jensen's inequality to minimize the objective function (logarithmic marginal likelihood) and then to maximize it. More generally, there is the MM algorithm~\cite{hunter:mm}, which uses not only Jensen's inequality but also Cauchy--Schwartz's inequality, arithmetic-geometric mean, or quadratic approximation to minimize and maximize. The MM algorithm is a broader class of meta-algorithm that includes the EM algorithm as a special case, and is expected to have a similar geometric structure, but its unified treatment as an algorithm on statistical manifolds is not obvious, and future research on this issue is expected.

\section*{Acknowledgments}
Part of this work is supported by JSPS KAKENHI No.JP22H03653 and JP22486199. 

 \newcommand{\noop}[1]{}


\begin{thebibliography}{10}

\bibitem{DEMP1977}
A.~P. Dempster, N.~M. Laird, and D.~B. Rubin.
\newblock Maximum likelihood from incomplete data via the {EM} algorithm.
\newblock {\em Journal of the Royal Statistical Society: Series B}, 39:1--38,
  1977.

\bibitem{10.1214/aos/1176346060}
C.~F.~Jeff Wu.
\newblock {On the Convergence Properties of the EM Algorithm}.
\newblock {\em The Annals of Statistics}, 11(1):95 -- 103, 1983.

\bibitem{10.2307/2337198}
Xiao-Li Meng and Donald~B. Rubin.
\newblock Maximum likelihood estimation via the {ECM} algorithm: A general
  framework.
\newblock {\em Biometrika}, 80(2):267--278, 1993.

\bibitem{1574231875723835008}
Imre Csisz\'ar and G\'abor Tusn\'ady.
\newblock Information geometry and alternating minimization procedures.
\newblock {\em Statistics and Decisions}, 1:205--237, 1984.

\bibitem{BA85746989}
Geoffrey~J. McLachlan and Thriyambakam Krishnan.
\newblock {\em The EM algorithm and extensions}.
\newblock Wiley series in probability and mathematical statistics.
  Wiley-Interscience, 2nd ed edition, 2008.

\bibitem{Balakrishnan2017}
Sivaraman Balakrishnan, Martin~J. Wainwright, and Bin Yu.
\newblock {Statistical guarantees for the EM algorithm: From population to
  sample-based analysis}.
\newblock {\em https://doi.org/10.1214/16-AOS1435}, 45(1):77--120, feb 2017.

\bibitem{DBLP:conf/aistats/KwonHC21}
Jeongyeol Kwon, Nhat Ho, and Constantine Caramanis.
\newblock On the minimax optimality of the {EM} algorithm for learning
  two-component mixed linear regression.
\newblock In {\em The 24th International Conference on Artificial Intelligence
  and Statistics, {AISTATS} 2021, April 13-15, 2021, Virtual Event}, pages
  1405--1413, 2021.

\bibitem{doi:10.1177/096228029700600105}
Jim Kay.
\newblock The {EM} algorithm in medical imaging.
\newblock {\em Statistical Methods in Medical Research}, 6(1):55--75, 1997.
\newblock PMID: 9185290.

\bibitem{McLachlan1997TheIO}
Geoffrey~J. McLachlan.
\newblock {The impact of the EM algorithm on medical statistics.}
\newblock {\em Statistical methods in medical research}, 6:1--2, 1997.

\bibitem{doi:10.1177/096228029700600103}
Niels~G Becker.
\newblock {Uses of the EM algorithm in the analysis of data on HIV/AIDS and
  other infectious diseases}.
\newblock {\em Statistical Methods in Medical Research}, 6(1):24--37, 1997.
\newblock PMID: 9185288.

\bibitem{Enders2003UsingTE}
Craig~K. Enders.
\newblock Using the expectation maximization algorithm to estimate coefficient
  alpha for scales with item-level missing data.
\newblock {\em Psychological methods}, 8:322--37, 2003.

\bibitem{DBLP:journals/tifs/MurakamiKH17}
Takao Murakami, Atsunori Kanemura, and Hideitsu Hino.
\newblock Group sparsity tensor factorization for re-identification of open
  mobility traces.
\newblock {\em {IEEE} Trans. Inf. Forensics Secur.}, 12(3):689--704, 2017.

\bibitem{DBLP:journals/popets/MurakamiHS18}
Takao Murakami, Hideitsu Hino, and Jun Sakuma.
\newblock Toward distribution estimation under local differential privacy with
  small samples.
\newblock {\em Proc. Priv. Enhancing Technol.}, 2018(3):84--104, 2018.

\bibitem{DBLP:journals/nn/IwasakiHTAM18}
Taishi Iwasaki, Hideitsu Hino, Masami Tatsuno, Shotaro Akaho, and Noboru
  Murata.
\newblock Estimation of neural connections from partially observed neural
  spikes.
\newblock {\em Neural Networks}, 108:172--191, 2018.

\bibitem{RUUD1991305}
Paul~A. Ruud.
\newblock Extensions of estimation methods using the {EM} algorithm.
\newblock {\em Journal of Econometrics}, 49(3):305--341, 1991.

\bibitem{10.1214/aoms/1177699147}
Leonard~E. Baum and Ted Petrie.
\newblock {Statistical Inference for Probabilistic Functions of Finite State
  Markov Chains}.
\newblock {\em The Annals of Mathematical Statistics}, 37(6):1554 -- 1563,
  1966.

\bibitem{amari2000methods}
S.~Amari and H.~Nagaoka.
\newblock {\em Methods of Information Geometry}.
\newblock Translations of mathematical monographs. American Mathematical
  Society, 2000.

\bibitem{10.2307/2337602}
Fumiyasu Komaki.
\newblock On asymptotic properties of predictive distributions.
\newblock {\em Biometrika}, 83(2):299--313, 1996.

\bibitem{bj/1178291931}
{Shun-ichi} Amari and Motoaki Kawanabe.
\newblock {Information geometry of estimating functions in semi-parametric
  statistical models}.
\newblock {\em Bernoulli}, 3(1):29 -- 54, 1997.

\bibitem{930911}
{Shun-ichi} Amari.
\newblock Information geometry on hierarchy of probability distributions.
\newblock {\em IEEE Transactions on Information Theory}, 47(5):1701--1711,
  2001.

\bibitem{Breiman1996}
Leo Breiman.
\newblock {Bagging predictors}.
\newblock {\em Machine Learning}, 24(2):123--140, 1996.

\bibitem{FREUND1997119}
Yoav Freund and Robert~E Schapire.
\newblock A decision-theoretic generalization of on-line learning and an
  application to boosting.
\newblock {\em Journal of Computer and System Sciences}, 55(1):119--139, 1997.

\bibitem{Baggins2004}
Tadayoshi Fushiki, Fumiyasu Komaki, and Kazuyuki Aihara.
\newblock On parametric bootstrapping and bayesian prediction.
\newblock {\em Scandinavian Journal of Statistics}, 31(3):403--416, 2004.

\bibitem{NIPS2001_71e09b16}
Guy Lebanon and John Lafferty.
\newblock Boosting and maximum likelihood for exponential models.
\newblock In T.~Dietterich, S.~Becker, and Z.~Ghahramani, editors, {\em
  Advances in Neural Information Processing Systems}, volume~14. MIT Press,
  2001.

\bibitem{DBLP:journals/neco/MurataTKE04}
Noboru Murata, Takashi Takenouchi, Takafumi Kanamori, and Shinto Eguchi.
\newblock {Information Geometry of U-Boost and Bregman Divergence}.
\newblock {\em Neural Comput.}, 16(7):1437--1481, 2004.

\bibitem{DBLP:journals/neco/TakenouchiEMK08}
Takashi Takenouchi, Shinto Eguchi, Noboru Murata, and Takafumi Kanamori.
\newblock Robust boosting algorithm against mislabeling in multiclass problems.
\newblock {\em Neural Comput.}, 20(6):1596--1630, 2008.

\bibitem{AMARI19951379}
{Shun-ichi} Amari.
\newblock Information geometry of the {EM} and em algorithms for neural
  networks.
\newblock {\em Neural Networks}, 8(9):1379--1408, 1995.

\bibitem{ay2017information}
Nihat Ay, J{\"u}rgen Jost, H{\^o}ng V{\^a}n~L{\^e}, and Lorenz
  Schwachh{\"o}fer.
\newblock {\em Information geometry}, volume~64.
\newblock Springer, 2017.

\bibitem{kobayashi1996foundations}
S.~Kobayashi and K.~Nomizu.
\newblock {\em Foundations of Differential Geometry, Volume 2}.
\newblock A Wiley Publication in Applied Statistics. Wiley, 1996.

\bibitem{BREGMAN1967200}
L.M. Bregman.
\newblock The relaxation method of finding the common point of convex sets and
  its application to the solution of problems in convex programming.
\newblock {\em USSR Computational Mathematics and Mathematical Physics},
  7(3):200--217, 1967.

\bibitem{FujimotoMurataEM2007}
Yu~Fujimoto and Noboru Murata.
\newblock A modified {EM} algorithm for mixture models based on {Bregman}
  divergence.
\newblock {\em Annals of the Institute of Statistical Mathematics}, 59:3--25,
  2007.

\bibitem{HE_IG_2022}
Hideitsu Hino and Shinto Eguchi.
\newblock Active learning by query by committee with robust divergences.
\newblock {\em Information Geometry}, \noop{3001}under review.

\bibitem{DBLP:journals/corr/abs-2201-02447}
Masahito Hayashi.
\newblock Bregman divergence based em algorithm and its application to
  classical and quantum rate distortion theory.
\newblock {\em CoRR}, abs/2201.02447, 2022.

\bibitem{1054753}
S.~Arimoto.
\newblock An algorithm for computing the capacity of arbitrary discrete
  memoryless channels.
\newblock {\em IEEE Transactions on Information Theory}, 18(1):14--20, 1972.

\bibitem{yeung2008information}
R.W. Yeung.
\newblock {\em Information Theory and Network Coding}.
\newblock Information Technology: Transmission, Processing and Storage.
  Springer US, 2008.

\bibitem{Toyota2020}
Shoji Toyota.
\newblock Geometry of arimoto algorithm.
\newblock {\em Information Geometry}, 3:183--198, 2020.

\bibitem{IkedaTA2004}
Shiro Ikeda, Toshiyuki Tanaka, and Shun ichi Amari.
\newblock Information geometry of turbo codes and low-density parity-check
  codes.
\newblock {\em IEEE Transaction on Information Theory}, 50:1097--1114, 2004.

\bibitem{r.a.bradley52:_rank_analy_of_incom_block_desig}
R.~A. Bradley and M.~Terry.
\newblock The rank analysis of incomplete block designs: {I}. the method of
  paired comparisons.
\newblock {\em Biometrika}, 39:324--345, 1952.

\bibitem{10.1214/aos/1028144844}
Trevor Hastie and Robert Tibshirani.
\newblock {Classification by pairwise coupling}.
\newblock {\em The Annals of Statistics}, 26(2):451 -- 471, 1998.

\bibitem{NIPS2004_825f9cd5}
Tzu-kuo Huang, Chih-jen Lin, and Ruby Weng.
\newblock {A Generalized Bradley-Terry Model: From Group Competition to
  Individual Skill}.
\newblock In L.~Saul, Y.~Weiss, and L.~Bottou, editors, {\em Advances in Neural
  Information Processing Systems}, volume~17. MIT Press, 2004.

\bibitem{DBLP:journals/neco/FujimotoHM11}
Yu~Fujimoto, Hideitsu Hino, and Noboru Murata.
\newblock {An Estimation of Generalized Bradley-Terry Models Based on the
  \emph{em} Algorithm}.
\newblock {\em Neural Comput.}, 23(6):1623--1659, 2011.

\bibitem{Plackett1975}
R.~L. Plackett.
\newblock The analysis of permutations.
\newblock {\em Applied Statistics}, 24(2):193--202, 1975.

\bibitem{DBLP:conf/pakdd/HinoFM09}
Hideitsu Hino, Yu~Fujimoto, and Noboru Murata.
\newblock Item preference parameters from grouped ranking observations.
\newblock In {\em Advances in Knowledge Discovery and Data Mining, 13th
  Pacific-Asia Conference, {PAKDD} 2009, Bangkok, Thailand, April 27-30, 2009,
  Proceedings}, pages 875--882, 2009.

\bibitem{DBLP:journals/neco/HinoFM10}
Hideitsu Hino, Yu~Fujimoto, and Noboru Murata.
\newblock A grouped ranking model for item preference parameter.
\newblock {\em Neural Comput.}, 22(9):2417--2451, 2010.

\bibitem{DBLP:conf/ic3k/FujimotoHM09}
Yu~Fujimoto, Hideitsu Hino, and Noboru Murata.
\newblock Item-user preference mapping with mixture models - data visualization
  for item preference.
\newblock In {\em {KDIR} 2009 - Proceedings of the International Conference on
  Knowledge Discovery and Information Retrieval, Funchal - Madeira, Portugal,
  October 6-8, 2009}, pages 105--111, 2009.

\bibitem{DBLP:journals/neco/SandoH20}
Keishi Sando and Hideitsu Hino.
\newblock Modal principal component analysis.
\newblock {\em Neural Comput.}, 32(10):1901--1935, 2020.

\bibitem{Lee1989}
Myoung.~J Lee.
\newblock Mode regression.
\newblock {\em Journal of Econometrics}, 42(3):337--349, 1989.

\bibitem{Hampel1986}
Frank~R. Hampel, Elvezio~M. Ronchetti, Peter~J. Rousseeuw, and Werner~A.
  Stahel.
\newblock {\em Robust Statistics - The Approach Based on Influence Functions}.
\newblock Wiley, 1986.

\bibitem{Huber2011}
Peter~J. Huber and Elvezio~M. Ronchetti.
\newblock {\em Robust Statistics}.
\newblock Wiley, 2011.

\bibitem{Pistone1995}
Giovanni Pistone and Carlo Sempi.
\newblock An infinite--dimensional geometric structure on the space of all the
  probability measures equivalent to a given one.
\newblock {\em Ann. Statist.}, 23(5):1543--1561, 10 1995.

\bibitem{Grasselli2010}
M.~R. Grasselli.
\newblock Dual connections in nonparametric classical information geometry.
\newblock {\em Annals of the Institute of Statistical Mathematics},
  62(5):873--896, Oct 2010.

\bibitem{Zhang2013}
Jun Zhang.
\newblock Nonparametric information geometry: From divergence function to
  referential-representational biduality on statistical manifolds.
\newblock {\em Entropy}, 15(12):5384--5418, 2013.

\bibitem{DBLP:conf/iconip/SandoAMH18}
Keishi Sando, Shotaro Akaho, Noboru Murata, and Hideitsu Hino.
\newblock Information geometric perspective of modal linear regression.
\newblock In {\em Neural Information Processing - 25th International
  Conference, {ICONIP} 2018, Siem Reap, Cambodia, December 13-16, 2018,
  Proceedings, Part {III}}, pages 535--545, 2018.

\bibitem{Sando2019}
Keishi Sando, Shotaro Akaho, Noboru Murata, and Hideitsu Hino.
\newblock {Information geometry of modal linear regression}.
\newblock {\em Information Geometry}, 2(1):43--75, jun 2019.

\bibitem{Yao2012}
Weixin Yao, Bruce~G Lindsay, and Runze Li.
\newblock Local modal regression.
\newblock {\em Journal of nonparametric statistics}, 24(3):647--663, 2012.

\bibitem{Gordon2012}
Gordon~C.R. Kemp and J.M.C.~Santos Silva.
\newblock Regression towards the mode.
\newblock {\em Journal of Econometrics}, 170(1):92 -- 101, 2012.

\bibitem{143375}
W.~Byrne.
\newblock Alternating minimization and boltzmann machine learning.
\newblock {\em IEEE Transactions on Neural Networks}, 3(4):612--620, 1992.

\bibitem{125867}
S.~Amari, K.~Kurata, and H.~Nagaoka.
\newblock Information geometry of boltzmann machines.
\newblock {\em IEEE Transactions on Neural Networks}, 3(2):260--271, 1992.

\bibitem{FUJIWARA1995317}
Akio Fujiwara and Shun ichi Amari.
\newblock Gradient systems in view of information geometry.
\newblock {\em Physica D: Nonlinear Phenomena}, 80(3):317--327, 1995.

\bibitem{NIPS1998_0771fc6f}
Shiro Ikeda, Shun-ichi Amari, and Hiroyuki Nakahara.
\newblock Convergence of the wake-sleep algorithm.
\newblock In M.~Kearns, S.~Solla, and D.~Cohn, editors, {\em Advances in Neural
  Information Processing Systems}, volume~11. MIT Press, 1998.

\bibitem{DBLP:journals/tmi/FletcherLPJ04}
P.~Thomas Fletcher, Conglin Lu, Stephen~M. Pizer, and Sarang~C. Joshi.
\newblock Principal geodesic analysis for the study of nonlinear statistics of
  shape.
\newblock {\em {IEEE} Trans. Medical Imaging}, 23(8):995--1005, 2004.

\bibitem{zhuang2020comprehensive}
Fuzhen Zhuang, Zhiyuan Qi, Keyu Duan, Dongbo Xi, Yongchun Zhu, Hengshu Zhu, Hui
  Xiong, and Qing He.
\newblock A comprehensive survey on transfer learning.
\newblock {\em Proceedings of the IEEE}, 109(1):43--76, 2020.

\bibitem{pan2008transfer}
Sinno~Jialin Pan, James~T Kwok, Qiang Yang, et~al.
\newblock Transfer learning via dimensionality reduction.
\newblock In {\em AAAI}, volume~8, pages 677--682, 2008.

\bibitem{DBLP:journals/neco/TakanoHAM16}
Ken Takano, Hideitsu Hino, Shotaro Akaho, and Noboru Murata.
\newblock Nonparametric \emph{e}-mixture estimation.
\newblock {\em Neural Comput.}, 28(12):2687--2725, 2016.

\bibitem{murata2009bregman}
Noboru Murata and Yu~Fujimoto.
\newblock Bregman divergence and density integration.
\newblock 2009.

\bibitem{hino2013information}
Hideitsu Hino and Noboru Murata.
\newblock Information estimators for weighted observations.
\newblock {\em Neural Networks}, 46:260--275, 2013.

\bibitem{hino2015non}
Hideitsu Hino, Kensuke Koshijima, and Noboru Murata.
\newblock Non-parametric entropy estimators based on simple linear regression.
\newblock {\em Computational Statistics \& Data Analysis}, 89:72--84, 2015.

\bibitem{DBLP:conf/iconip/AkahoHM19}
Shotaro Akaho, Hideitsu Hino, and Noboru Murata.
\newblock On a convergence property of a geometrical algorithm for statistical
  manifolds.
\newblock In {\em Neural Information Processing - 26th International
  Conference, {ICONIP} 2019, Sydney, NSW, Australia, December 12-15, 2019,
  Proceedings, Part {V}}, pages 262--272, 2019.

\bibitem{akaho2004pca}
Shotaro Akaho.
\newblock {The e-PCA and m-PCA: Dimension reduction of parameters by
  information geometry}.
\newblock In {\em 2004 IEEE International Joint Conference on Neural Networks
  (IEEE Cat. No. 04CH37541)}, volume~1, pages 129--134. IEEE, 2004.

\bibitem{collins2001generalization}
Michael Collins, Sanjoy Dasgupta, and Robert~E Schapire.
\newblock A generalization of principal components analysis to the exponential
  family.
\newblock {\em Advances in neural information processing systems}, 14, 2001.

\bibitem{watanabe2009variational}
Kazuho Watanabe, Shotaro Akaho, Shinichiro Omachi, and Masato Okada.
\newblock Variational bayesian mixture model on a subspace of exponential
  family distributions.
\newblock {\em IEEE transactions on neural networks}, 20(11):1783--1796, 2009.

\bibitem{lee2000algorithms}
Daniel Lee and H~Sebastian Seung.
\newblock Algorithms for non-negative matrix factorization.
\newblock {\em Advances in neural information processing systems}, 13, 2000.

\bibitem{cichocki2009nonnegative}
A.~Cichocki, R.~Zdunek, A.H. Phan, and S.~Amari.
\newblock {\em Nonnegative Matrix and Tensor Factorizations: Applications to
  Exploratory Multi-way Data Analysis and Blind Source Separation}.
\newblock Wiley, 2009.

\bibitem{akaho2018geometrical}
Shotaro Akaho, Hideitsu Hino, Neneka Nara, and Noboru Murata.
\newblock Geometrical formulation of the nonnegative matrix factorization.
\newblock In {\em International Conference on Neural Information Processing},
  pages 525--534. Springer, 2018.

\bibitem{ishibashi2022principal}
Hideaki Ishibashi and Shotaro Akaho.
\newblock Principal component analysis for {Gaussian} process posteriors.
\newblock {\em Neural Computation}, 34(5):1189--1219, 2022.

\bibitem{akaho2008dimension}
Shotaro Akaho.
\newblock Dimension reduction for mixtures of exponential families.
\newblock In {\em International Conference on Artificial Neural Networks},
  pages 1--10. Springer, 2008.

\bibitem{NIPS2014_5ca3e9b1}
Ian Goodfellow, Jean Pouget-Abadie, Mehdi Mirza, Bing Xu, David Warde-Farley,
  Sherjil Ozair, Aaron Courville, and Yoshua Bengio.
\newblock Generative adversarial nets.
\newblock In Z.~Ghahramani, M.~Welling, C.~Cortes, N.~Lawrence, and K.Q.
  Weinberger, editors, {\em Advances in Neural Information Processing Systems},
  volume~27. Curran Associates, Inc., 2014.

\bibitem{5605355}
XuanLong Nguyen, Martin~J. Wainwright, and Michael~I. Jordan.
\newblock Estimating divergence functionals and the likelihood ratio by convex
  risk minimization.
\newblock {\em IEEE Transactions on Information Theory}, 56(11):5847--5861,
  2010.

\bibitem{NIPS2016_cedebb6e}
Sebastian Nowozin, Botond Cseke, and Ryota Tomioka.
\newblock f-gan: Training generative neural samplers using variational
  divergence minimization.
\newblock In D.~Lee, M.~Sugiyama, U.~Luxburg, I.~Guyon, and R.~Garnett,
  editors, {\em Advances in Neural Information Processing Systems}, volume~29.
  Curran Associates, Inc., 2016.

\bibitem{NIPS2017_2f2b2656}
Richard Nock, Zac Cranko, Aditya~K Menon, Lizhen Qu, and Robert~C Williamson.
\newblock f-gans in an information geometric nutshell.
\newblock In I.~Guyon, U.~Von Luxburg, S.~Bengio, H.~Wallach, R.~Fergus,
  S.~Vishwanathan, and R.~Garnett, editors, {\em Advances in Neural Information
  Processing Systems}, volume~30. Curran Associates, Inc., 2017.

\bibitem{JMLR:v13:gretton12a}
Arthur Gretton, Karsten~M. Borgwardt, Malte~J. Rasch, Bernhard Sch{{\"o}}lkopf,
  and Alexander Smola.
\newblock A kernel two-sample test.
\newblock {\em Journal of Machine Learning Research}, 13(25):723--773, 2012.

\bibitem{pmlr-v37-li15}
Yujia Li, Kevin Swersky, and Rich Zemel.
\newblock Generative moment matching networks.
\newblock In Francis Bach and David Blei, editors, {\em Proceedings of the 32nd
  International Conference on Machine Learning}, volume~37 of {\em Proceedings
  of Machine Learning Research}, pages 1718--1727, Lille, France, 07--09 Jul
  2015. PMLR.

\bibitem{10.5555/3020847.3020875}
Gintare~Karolina Dziugaite, Daniel~M. Roy, and Zoubin Ghahramani.
\newblock Training generative neural networks via maximum mean discrepancy
  optimization.
\newblock In {\em Proceedings of the Thirty-First Conference on Uncertainty in
  Artificial Intelligence}, UAI'15, page 258^^e2^^80^^93267, Arlington,
  Virginia, USA, 2015. AUAI Press.

\bibitem{pmlr-v70-arjovsky17a}
Martin Arjovsky, Soumith Chintala, and L{\'e}on Bottou.
\newblock {W}asserstein generative adversarial networks.
\newblock In Doina Precup and Yee~Whye Teh, editors, {\em Proceedings of the
  34th International Conference on Machine Learning}, volume~70 of {\em
  Proceedings of Machine Learning Research}, pages 214--223. PMLR, 06--11 Aug
  2017.

\bibitem{hunter:mm}
Hunter DR and Lange K.
\newblock A tutorial on {MM} algorithms.
\newblock {\em The American Statistician}, (58), 2004.

\end{thebibliography}
\end{document}